\documentclass{article} 
\usepackage{float}
\usepackage[table,xcdraw]{xcolor}
\usepackage[final]{colm2025_conference}

\usepackage{microtype}
\usepackage{hyperref}
\usepackage{url}
\usepackage{booktabs}
\usepackage{graphicx}
\usepackage{multirow}
\usepackage{multicol}
\usepackage{subcaption}
\usepackage[utf8]{inputenc}
\usepackage{graphicx} 
\usepackage{wrapfig}
\usepackage{booktabs} 
\usepackage{lipsum}
\usepackage{tikz}        
\usepackage{amsmath}
\usepackage{capt-of}
\usepackage{makecell}
\usepackage{amsthm}
\usepackage{textcomp}

\DeclareMathOperator*{\argmin}{arg\,min}

\newcommand{\circleGreen}{\tikz\draw[fill=green,draw=green] (0,0) circle (2pt);}
\newcommand{\circleYellow}{\tikz\draw[fill=yellow,draw=yellow] (0,0) circle (2pt);}
\newcommand{\circleRed}{\tikz\draw[fill=red,draw=red] (0,0) circle (2pt);}

\definecolor{darkblue}{rgb}{0, 0, 0.5}
\hypersetup{colorlinks=true, citecolor=darkblue, linkcolor=darkblue, urlcolor=darkblue}

\usepackage{amsmath} 
\newtheorem{theorem}{Theorem}

\title{M²IV: Towards Efficient and Fine-grained Multimodal \\ In-Context Learning via Representation Engineering}

\author{Yanshu Li$^{1}$\thanks{Equal contributions.}\hspace{1em}
  Yi Cao$^{2}$\footnotemark[1]\hspace{1em}
  Hongyang He$^{3}$\hspace{1em}
  Qisen Cheng$^{4}$\hspace{1em}
  Xiang Fu$^{5}$\hspace{1em}\\
  \textbf{Xi Xiao}$^{6}$\hspace{1em}
  \textbf{Tianyang Wang}$^{6}$\hspace{1em}
  \textbf{Ruixiang Tang}$^{7}$\thanks{Corresponding author: ruixiang.tang@rutgers.edu}
  \\
$^{1}$Brown University  $^{2}$Soochow University $^{3}$University of Warwick 
 $^{4}$Samsung US  \\ $^{5}$Boston University $^{6}$University of Alabama at Birmingham
$^{7}$Rutgers University\\
}

\usepackage{float}

\begin{document}

\ifcolmsubmission
\linenumbers
\fi

\maketitle
\begin{abstract}
Multimodal in-context learning (ICL) equips Large Vision-language Models (LVLMs) with the ability to adapt to new tasks via multiple user-provided demonstrations, without requiring any model parameter updates. However, its effectiveness is constrained by the token-intensive nature of multimodal inputs and the complexity of cross-modal few-shot reasoning, which together hinder LVLMs from extracting useful patterns from demonstrations. To address these challenges, we propose \textbf{M²IV}, a novel representation engineering approach that replaces explicit token-level demonstrations with a set of learnable Multimodal In-context Vectors directly injected into the residual streams of LVLMs. By analyzing the distinct roles of multi-head attention (MHA) and multi-layer perceptrons (MLP) in the ICL process, we design a training strategy that enables M²IV to perform fine-grained semantic distillation and robust cross-modal representation learning. M²IV not only improves performance across diverse tasks and LVLMs but also significantly reduces token overhead, enabling graceful scaling to many-shot scenarios. To further enhance usability, we introduce \textbf{VLibrary}, a repository that stores trained M²IVs for flexible retrieval and injection. With VLibrary, users can steer pre-trained LVLMs in a customized manner that meets diverse requirements. Extensive experiments demonstrate that M²IV consistently outperforms vanilla ICL and prior representation engineering baselines, achieving an average accuracy gain of 3.74\% with substantial improvements in overall efficiency.

\end{abstract}
\section{Introduction}
\label{intro}

Large Vision-language Models (LVLMs) unify visual and textual modalities within the representation space of their Large Language Model (LLM) backbones, thereby gaining advanced multimodal understanding and generation capabilities \citep{zhou2022learning, zhang2024vision, laurenccon2025matters}. They are being deployed in an ever-growing array of vision–language (VL) applications \citep{health, e-commerce}. As task complexity rises, effectively and efficiently guiding LVLMs to adapt to new tasks becomes increasingly important. In-context learning (ICL) offers a promising solution that allows models to quickly learn from demonstrations directly inserted into the prompts, without updating any parameters \citep{ICL1, ICL2}.

Despite recent progress in applying multimodal ICL to tasks such as image captioning and classification \citep{captioning, class}, extending it to more complex or knowledge-intensive tasks remains a fundamental challenge \citep{adv}. This difficulty stems from two key limitations. First, the token-heavy nature of interleaved image-text input makes it hard to incorporate external knowledge efficiently. A common four- or eight-shot prompt introduces far greater inference latency than a single-image input. When more supporting knowledge is needed \citep{RAG}, users may have to split the prompt into several segments, as most LVLMs still provide limited context windows. This further increases latency and can trigger catastrophic forgetting \citep{forget}. Second, ICL can be surprisingly unstable due to its high sensitivity to few-shot prompt designs. Even state-of-the-art (SOTA) models are found to exhibit substantial performance fluctuations depending on the format, content, and order of the demonstrations provided \citep{dpsk}. These challenges are further amplified in multimodal scenarios, where aligning visual and textual features adds another layer of complexity \citep{sensitive2, sensitive}.

To address these issues, representation engineering has emerged as a feasible direction \citep{persona}. This line of work views ICL as shifts applied to the model’s internal activations \citep{shift}. It extracts these ICL-induced shifts as vectors and reinjects them during zero-shot inference to implicitly encode task-specific guidance. For instance, \citet{TV} derive task vectors from the hidden state of the context's last token, while \citet{FV} compute function vectors by averaging the outputs of key attention heads. Although these methods perform well on simple tasks with clear mappings, they fall short on complex multimodal tasks like visual question answering (VQA). Consequently, \citet{live} propose training in-context vectors to capture richer features. However, their approach still suffers from rapid degradation as complexity increases. This exposes a shared limitation: an inability to effectively model intricate interactions in multimodal ICL. To overcome this shortcoming, two key challenges must be tackled simultaneously: \textbf{I.} identifying the critical task mappings within individual demonstrations and the interdependencies among different demonstrations, and \textbf{II.} achieving fine-grained semantic distillation of the complex multimodal information.

\begin{figure}[t]
\begin{center}
\includegraphics[width=\textwidth]{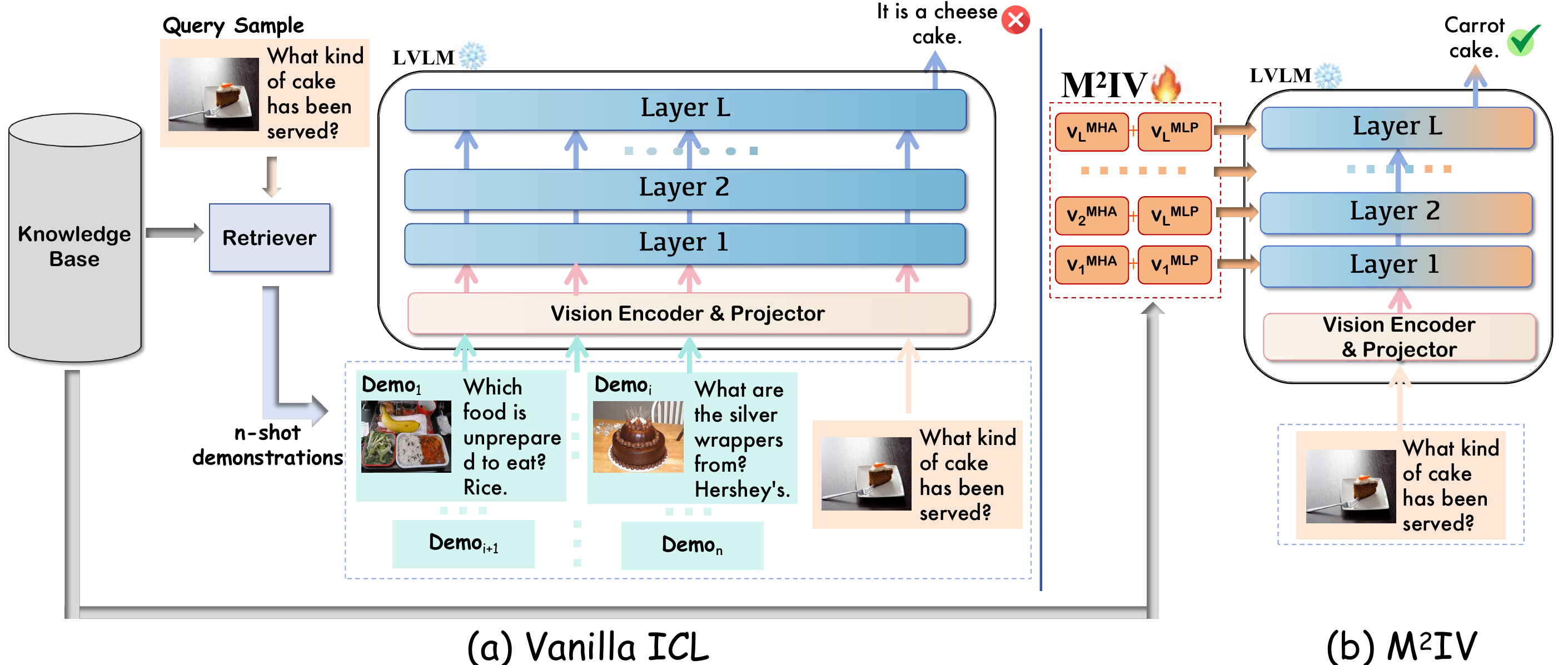}
\end{center}
\vspace{-10pt}
\caption{(a) In Vanilla ICL, for a given query sample, we retrieve $n$ instances from a knowledge base as demonstrations and feed them together with the query sample into the LVLM. (b) In contrast, we train a set of vectors emulating $n$-shot ICL using the same knowledge base and inject these vectors into the LVLM, greatly reducing token consumption and boosting model performance.}
\label{teaser}
\vspace{-20pt}
\end{figure}

Building on these insights, we introduce M²IV, a novel method that harnesses the distinct roles of multi-head attention (MHA) and multi-layer perceptrons (MLP) in multimodal ICL. M²IV assigns a set of learnable vectors to the MHA branch and another set to the MLP branch at each decoder layer, as illustrated in Figure \ref{teaser}(b). We design a dedicated training strategy that enables these vectors to absorb the deep semantic patterns captured by MHA in the demonstrations and to simulate the distilled information produced by MLP. The experiments show that M²IV achieves SOTA performance in 18 of 21 experiments conducted on three LVLMs and seven benchmarks, while utilizing only about 24\% of the total training data used by LIVE. Furthermore, M²IV delivers an efficiency breakthrough as the reduction in inference time fully compensates for the one-time training cost and yields even greater cost-effectiveness as adoption scales. To further explore the potential of M²IV, we present VLibrary, a container that stores trained M²IVs and supports on-demand retrieval for plug-and-play use. VLibrary can be seamlessly incorporated into real-world systems and applied to address critical challenges in the LVLM domain. 

Our main contributions are: (1) We analyze the unique roles of MHA and MLP in multimodal ICL, revealing how MHA drives semantic integration and MLP refines fine-grained details. (2) We propose M²IV, which simultaneously achieves complex semantic understanding and fine-grained semantic distillation, demonstrating superior performance across three LVLMs and seven diverse benchmarks. (3) We introduce VLibrary and use it to address three critical challenges: cross-modal alignment, output customization, and safety, thereby offering a novel and promising pathway for future LVLM research.

\section{Background and Related Works}
\label{section2}

ICL emerges as a pivotal capability as language models scale \citep{rela1, rela2} and has since been extended to the multimodal domain. Several LVLMs are specifically pretrained or fine-tuned to attain multimodal ICL capability. Among the most notable examples are OpenFlamingo \citep{OF} and IDEFICS \citep{IDE}, which serve as key components of our work. With demand for multimodal ICL steadily increasing, supporting it within fixed context windows has become a critical feature of modern LVLMs, such as Qwen2.5VL \citep{qwen} and Grok3 \citep{grok3}.

In LVLMs, ICL is defined as the process in which a pretrained model $\mathcal{M}$ is provided with a prompt containing a multimodal context $\mathbf{C}$ (absent in zero-shot settings) and a query sample $\mathbf{Q}$. To generate the answer $\mathbf{A}$, the model computes the conditional probability:
\begin{equation} 
P_{\mathcal{M}}\left(\mathbf{A}|\mathbf{Q}, \mathbf{C}\right)
\end{equation}
via a feed-forward pass\footnote{We refer to this process as Vanilla ICL. In this work, we focus on image-to-text tasks.}. $\mathbf{C}$ includes few-shot demonstrations, as shown in Figure \ref{teaser}(a).

\begin{wrapfigure}{r}{0.6\textwidth}
  \centering
  \vspace{-10pt}
  \includegraphics[width=0.6\textwidth]{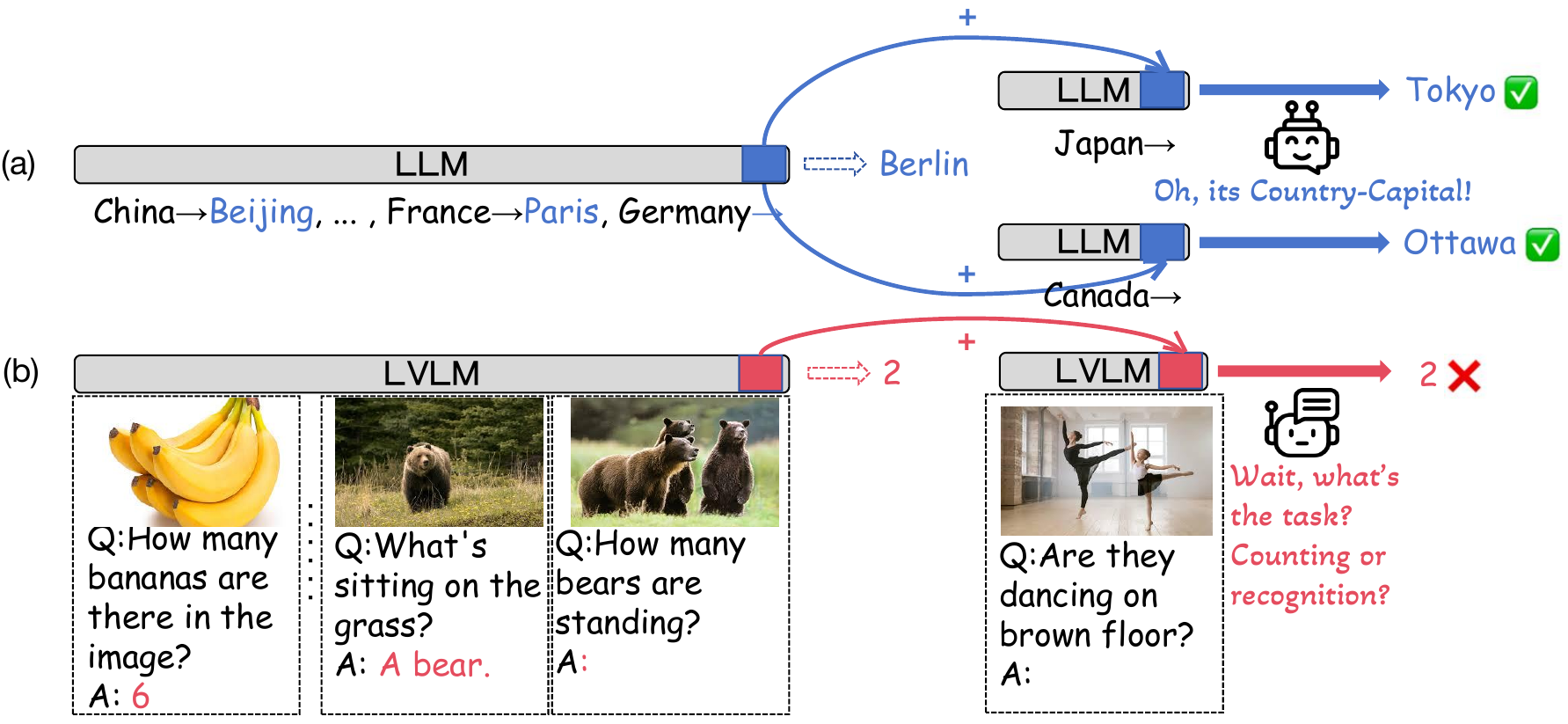} 
  \caption{Overview of applying Function Vector to (a) the Country-Capital task and (b) the VQA task, which requires complex reasoning. While it performs well in (a), it falls short of fully representing the context in (b).}
  \label{show}
  \vspace{-5pt}
\end{wrapfigure}

With the prevalence of ICL, researchers have been investigating its underlying mechanisms  \citep{rela3, rela4}, They point out that key attention dynamics, most notably skill neurons and induction heads \citep{rela5}, play a crucial role in its success. These insights have driven advances in \textbf{representation engineering} for ICL. Firstly, \citet{TV} and \citet{FV} nearly simultaneously propose Task Vector (TV) and Function Vector (FV). Building on this, \citet{ICV} obtain layer-wise In-Context Vectors (ICVs) by comparing the hidden states of the final tokens in the original and target sequences, then applying PCA to distill task-relevant information. Expanding ICV further, \citet{I2CL} propose Implicit ICL (I2CL), which extracts residual stream deltas at each demonstration’s last token position across layers and utilizes a set of coefficients to regulate their injection. As noted in §\ref{intro} and illustrated in Figure \ref{show}, these methods often underperform in multimodal ICL. \citet{live} attribute this gap to the static extraction-injection paradigm and propose an attention shift-based training method, LIVE, yet it still neglects the essential role of fine-grained semantic distillation.

\section{Methodology}
\begin{figure}[t]
\begin{center}
\includegraphics[width=\textwidth]{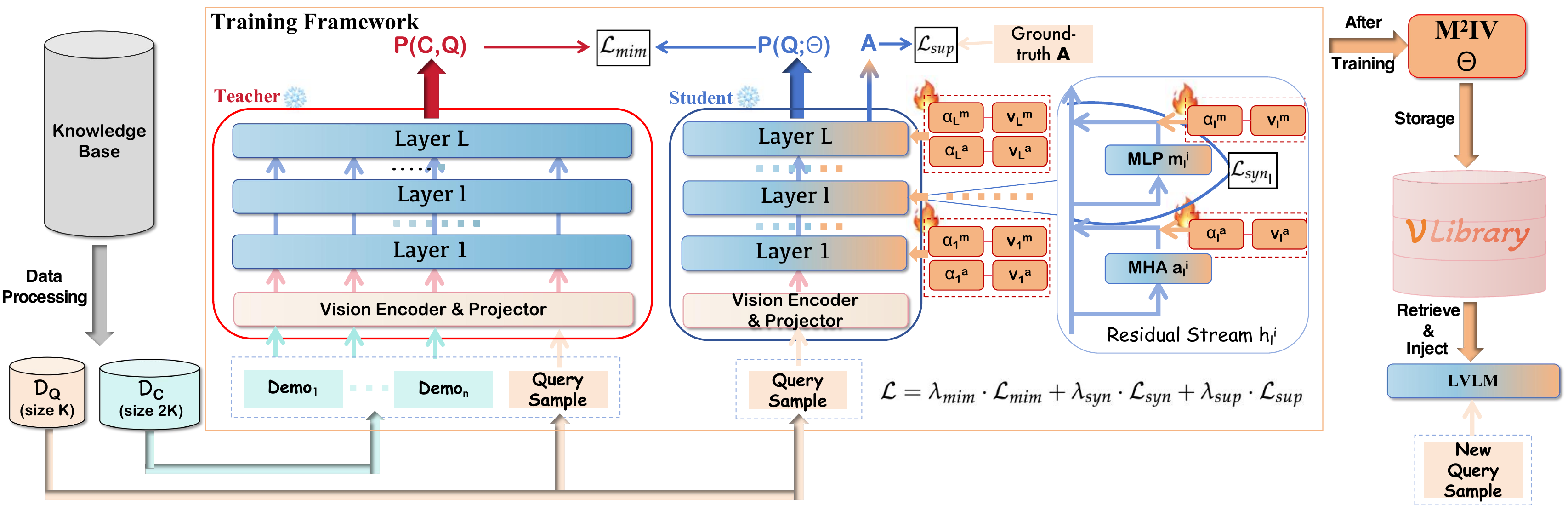}
\end{center}
\caption{Pipeline of M²IV's training and storage. We begin by constructing the training sets $\mathcal{D}_\textbf{Q}$ and $\mathcal{D}_\textbf{C}$. from an existing knowledge base. Next, a self-distillation framework employing three losses ($\mathcal{L}_{mim}$, $\mathcal{L}_{syn}$ and $\mathcal{L}_{sup}$) is used to train M²IV. Finally, the trained M²IV is stored in VLibrary for on-demand retrieval.}
\label{pipeline}
\vspace{-10pt}
\end{figure}

In this section, we detail the proposed M²IV framework by analyzing the roles of MHA and MLP in LVLM ICL, which motivates our definition of the M²IV parameters (§\ref{3.1}). We then present the complete training strategy (§\ref{3.2}). Finally, we introduce VLibrary, the storage for M²IVs (§\ref{3.3}). The overview pipeline is illustrated in Figure \ref{pipeline}.
\subsection{Anchoring M²IV}
\label{3.1}
For an $L$-layer LVLM $\mathcal{M}$ processing an input sequence of length $I$, the residual stream architecture is recursively defined as follows, with $l \in \left \{ 1, 2, \dots, L\right \} $ and $i \in \left \{ 1, 2, \dots, I\right \} $: 
\begin{align}
    \mathbf{a}_{l}^{i}&=\text{MHA}\left(\mathbf{h}_{l-1}^{i};\theta_{l}\right),
    \\
    \mathbf{m}_{l}^i&=\text{MLP}\left(\mathbf{h}_{l-1}^{i}+\mathbf{a}_{l}^{i};\mathbf{W}_{l}\right), 
    \\
    \mathbf{h}_{l}^{i}&=\mathbf{h}_{l-1}^{i}+\mathbf{a}_{l}^{i}+\mathbf{m}_{l}^{i},
\end{align}
where $\mathbf{h}_{l}^{i}\in \mathbb{R}^{d_\mathcal M}$ denotes the residual stream at layer $l$ and position $i$, with hidden dimension $d_\mathcal M$. The functions $\text{MHA}\left(\cdot; \theta_l\right): \mathbb{R}^{d_\mathcal{M}} \to \mathbb{R}^{d_\mathcal{M}}$ and $\text{MLP}\left(\cdot; \mathbf{W}_l\right): \mathbb{R}^{d_\mathcal{M}} \to \mathbb{R}^{d_\mathcal{M}}$ denote the MHA and MLP transformations at layer $l$, parameterized by $\theta_l$ and $\mathbf{W}_l$, respectively.

Previous research \citep{parts, I2CL} has demonstrated that MHA and MLP branches play distinct roles in ICL. Based on these insights, we infer that in multimodal ICL, MHA dynamically allocates attention to capture both the internal semantics within demonstrations and the interactions among them. MLP further extracts, filters, and stores key information in a fine-grained manner, conveying more aggregated yet nuanced features. To validate our inferences, we explore their computational invariance for a demonstration matrix $\mathbf{C}\in \mathbb{R}^{C \times d_{\mathcal{M}}}$ with $C$ tokens, and two invariance properties are established.
\begin{theorem}[MHA Computational Invariance] \label{mha}
There exists $\Psi:\mathbf{D}_{\Psi}\to \mathbb{R}^{d_{\mathcal{M}}}$ such that, for any query matrix $\mathbf{Q}\in \mathbb{R}^{I \times d_{\mathcal{M}}}$ and residual stream $\mathbf{h}^{i}\in \mathbb{R}^{d_{\mathcal{M}}}$, there exist $\zeta^{i}$, $\eta^{i}\in \mathbb{R}$ satisfying:
\begin{equation} \label{eq:core1}
    \mathrm{Attn}\left(
    \mathbf{h}^{i}, 
    \begin{bmatrix}
    \mathbf{C}^{\top}
    \:
    \mathbf{Q}^{\top}
    \end{bmatrix}^{\top}, \begin{bmatrix}
    \mathbf{C}^{\top}
    \:
    \mathbf{Q}^{\top}
    \end{bmatrix}^{\top}
    \right)
    =\zeta^{i}\cdot \Psi \left(\mathbf{h}^{i}, \mathbf{C}, \mathbf{C}\right)
    +
    \eta^{i}\cdot \mathrm{Attn}\left(\mathbf{h}^{i}, \mathbf{Q}, \mathbf{Q}\right),
\end{equation}
where $\mathrm{Attn}\left(\cdot\right)$ is a function denoting the self-attention mechanism. The query matrix $\mathbf{Q}$ is $\mathbf{Q}=\mathrm{concat}\left(\mathrm{Enc^{img}}\left(I\right),\mathrm{Enc^{txt}}\left(Q\right)\right)$, with $I$ and $Q$ being the query sample's image and question. Here, $\mathrm{Enc^{img}}\left(\cdot \right)$ and $\mathrm{Enc^{txt}}\left(\cdot \right)$ denote the encoding functions for visual and textual features, respectively.
\end{theorem}
\textit{Remark.} Theorem \ref{mha} shows that the self-attention mechanism decomposes into two parts: a query-only component, $\text{Attn}(\mathbf{h}^{i}, \mathbf{Q}, \mathbf{Q})$, and a context-augmented component, $\Psi(\mathbf{h}^{i}, \mathbf{C}, \mathbf{C})$. This decomposition highlights MHA's role in dynamically allocating attention, allowing $\mathcal{M}$ to integrate query-specific focus with contextual insights from demonstrations in ICL. 

\begin{theorem}[MLP Computational Invariance]\label{mlp}
For linear transformation matrix $\mathbf{W} \in \mathbb{R}^{d_{\mathcal{M}} \times d_{\mathcal{M}}}$ , there exists $\psi_{\mathbf{W}}: \mathbb{R}^{d_{\mathcal{M}}}\to \mathbb{R}^{d_{\mathcal{M}}}$ such that, for any token position $i$, there exist $\zeta^{i}$, $\eta^{i}\in \mathbb{R}$ satisfying:
    \begin{equation}
        \mathrm{MLP}\left(\mathbf{a}^i_{\mathbf{C},\mathbf{Q}}\right)=\zeta^i\cdot \psi_{\mathbf{W}}\left(\mathbf{a}^i_{\mathbf{C}}\right)+\eta^i\cdot \mathrm{MLP}\left(\mathbf{a}^i_{\mathbf{Q}}\right),
    \end{equation}
    where $\mathrm{MLP}\left(\cdot\right)$ represents the MLP mechanism, and $\mathbf{a}^i_{\mathbf{Q},\mathbf{C}}$, $\mathbf{a}^i_{\mathbf{Q}}$, and $\mathbf{a}^i_{\mathbf{C}}$ are the MHA outputs at the $i$-th token position with both query and context, with query-only, and with context-only, respectively.
\end{theorem} 
\textit{Remark.} Theorem \ref{mlp} shows that the MLP mechanism decomposes into two parts: a query-only component, $\text{MLP}(\mathbf{a}^i_{\mathbf{Q}})$, and a context-enhanced component, $\psi_\mathbf{W}(\mathbf{a}^i_{\mathbf{C}})$. Through weighted combinations, MLP extracts and retains key features from both query and context, enabling $\mathcal{M}$ to convey more aggregated yet nuanced representations in ICL. 

Theorems \ref{mha} and \ref{mlp}, with proofs provided in Appendices \ref{proof_mha} and \ref{proof_mlp}, illustrate the dual processing pathways inherent in multimodal ICL. This also offers a new perspective on the distinct roles of each layer in ICL, supplementing \citet{wang2023label}: shallow layers primarily aggregate information within the multimodal context $\mathbf{C}$; intermediate layers rely more on MHA to capture deeper semantic details; and deep layers tend to refine and integrate prior information via MLP. These insights drive our layer-wise design of M²IV.

Based on the analysis presented above, we propose M²IV for the fine-grained representation of multimodal ICL. M²IV assigns a learnable vector and a weight factor to both MHA and MLP branches at each layer of an LVLM. Specifically, we define:
\begin{align}
     \text{MHA: } & \mathbf{V}^a = \{\mathbf{v}^a_1, \mathbf{v}^a_2, \ldots, \mathbf{v}^a_L\}, \boldsymbol{\alpha}^a = \{\alpha_1^a, \alpha_2^a, \ldots, \alpha_L^a\}; \\
     \text{MLP: } & \mathbf{V}^m = \{\mathbf{v}^m_1, \mathbf{v}^m_2, \ldots, \mathbf{v}^m_L\},\boldsymbol{\alpha}^m = \{\alpha_1^m, \alpha_2^m, \ldots, \alpha_L^m\}.
\end{align}
Here, $\mathbf{v}_l^a, \mathbf{v}_l^m \in \mathbb{R}^{d_{\mathcal M}}$ and \(\alpha_l^a,\alpha_l^m \in \mathbb{R}\). The complete set of M²IV is represented by $\Theta=\left \{\alpha_{l}^{a}, \mathbf{v}_{l}^{a}, \alpha_{l}^{m}, \mathbf{v}_{l}^{m}\right \}^{L}_{l=1}$, which can be injected directly into the LVLM's residual streams. The updated residual stream is recursively defined for $l \in \left \{ 1,2,\dots,L \right \}$ and $i \in \left \{ 1,2,\dots,I \right \}$ as:
\begin{equation}
\label{injection}
    \mathbf{h}_{l}^{i}=\mathbf{h}_{l-1}^{i}+
    \left(\mathbf{a}_{l}^{i}+\alpha_{l}^{a}\cdot \mathbf{v}_{l}^{a}\right)+\left(\mathbf{m}_{l}^{i}+\alpha_{l}^{m}\cdot \mathbf{v}_{l}^{m}\right).
\end{equation}

\subsection{Training M²IV}
\label{3.2}
Given a dataset $\mathcal{D} = \{\left(I_j, Q_j,A_j\right)\}^{|\mathcal{D}|}_{j=1}$\footnote{$I_j$, $Q_j$ and $A_j$ denote the image, question and answer of the $j$-th instance in $\mathcal{D}$, respectively.} used for ICL, we aim to train M²IV to capture the effect of providing any $n$ instances from $\mathcal{D}$ as contexts under a specific retrieval strategy $\mathcal{R}$. 

We first process $\mathcal{D}$ as follows: (1) For each instance $\left(I_j, Q_j, A_j\right) \in \mathcal{D}$, we separately embed its $I_j$ and $Q_j$ with a CLIP model and then concatenate them to form a joint embedding as its semantic representation. (2) We cluster all joint embeddings via k-means with cosine similarity, partitioning $\mathcal{D}$ into $K$ clusters. From each cluster, we select the instance closest to the centroid (after removing its answer) to construct the query sample set $\mathcal{D}_{\mathbf{Q}}$ (size $K$); $\mathcal{D}  \setminus \mathcal{D}_\mathbf{Q}$ becomes the support set. (3) For each query sample in $\mathcal{D}_\mathbf{Q}$, we apply $\mathcal{R}$ to retrieve $n$ instances from the support set, forming an $n$-shot context. These $K$ contexts constitute the context set $\mathcal{D}_\mathbf{C}$. We augment $\mathcal{D}_\mathbf{C}$ by creating a copy of each context with its demonstrations randomly shuffled and adding it. Further details are provided in Appendix \ref{app:data}.

After obtaining the training data, we employ a self-distillation framework to train M²IV. The teacher model processes each $n$-shot context in $\mathcal{D}_\mathbf{C}$ with its corresponding query sample in $\mathcal{D}_\mathbf{Q}$ to perform Vanilla ICL. The student model is injected with the initialized $\Theta$ and receives only the same query sample as input. Based on this, we introduce \textbf{mimicry loss}:
\begin{equation}
\mathcal{L}_{mim}=\mathcal{T}^2\cdot \mathbb{D}_{KL}\left(P_{\mathcal{M}}^{\mathcal{T}}(\mathbf{C},\mathbf{Q})\parallel P_{\mathcal{M}}(\mathbf{Q};\Theta)\right),\quad \mathbf{C}\in \mathcal{D}_\mathbf{C},\mathbf{Q}\in \mathcal{D}_\mathbf{Q},
\end{equation}
where $\mathbb{D}_{KL}\left(\cdot \parallel \cdot\right)$ is the KL divergence. We apply temperature scaling with a parameter $\mathcal{T}$ to $P_{\mathcal{M}}\left(\mathbf{C},\mathbf{Q}\right)$ to facilitate smooth knowledge distillation and mitigate overconfidence.

Beyond distributional alignment, we seek to capitalize on the synergy between the MHA and MLP branches. Thus, we apply \textbf{synergistic loss} to the student model, which is designed to fortify their coherence and complementarity. Formally, we define it as follows:
\begin{equation}
\mathcal{L}_{syn} =   \sum_{l=1}^L\left(\sum_{i=1}^{d_\mathcal{M}} \left(1 - \mathbf{M}_{ii}^{l}\right)^2 + \gamma \cdot \sum_{i=1}^{d_\mathcal{M}} \sum_{j\ne i} {\mathbf{M}_{ij}^{l}}^2\right),
\end{equation}
where $\mathbf{M}^{l}=\left(\mathbf{Z}_{l}^{a}\left(\Theta\right)\right)^{\top}{\mathbf{Z}}_{l}^{m}\left(\Theta\right)\in \mathbb{R}^{d_\mathcal{M}\times d_\mathcal{M}}$ is a cross-view correlation matrix. $\mathbf{Z}_{l}^{\mathbf{a}}\left(\Theta\right)$ and $\mathbf{Z}_{l}^{\mathbf{m}}\left(\Theta\right)$ denote the normalized output of MHA and MLP at layer $l$ after injecting $\Theta$, respectively. $\mathbf{M}_{ij}^{l}$ denotes the element at the $i$-th row and $j$-th column of $\mathbf{M}^{l}$. The hyperparameter $\gamma$ balances consistency within dimensions and orthogonality across dimensions \footnote{The detailed computation process is presented in Appendix \ref{app:loss}.}.

Finally, we employ a standard cross-entropy loss function as \textbf{supervised loss}, ensuring that the student model’s predictions remain faithful to the answer $\mathbf{A}=\{\mathbf{A}_1,\mathbf{A}_2,...,\mathbf{A}_T\}$:
\begin{equation}
\mathcal{L}_{sup}=-   \sum_{t=1}^{T} \log P_{\mathcal{M}}\left(\mathbf{A}_{t} \mid  \mathbf{Q},\mathbf{A}_{:<t}; \Theta\right),\quad \mathbf{Q}\in \mathcal{D}_\mathbf{Q}.
\end{equation}

The final training objective combines these losses as a weighted sum:
\begin{equation}\label{eq:loss_final}
\mathcal{L}= \lambda_{mim} \cdot \mathcal{L}_{mim}+\lambda_{syn} \cdot \mathcal{L}_{syn}+\lambda_{sup} \cdot \mathcal{L}_{sup}
,
\end{equation}
where $\lambda_{mim}$, $\lambda_{syn}$, and $\lambda_{sup}\in \left(0, 1\right)$ denote the hyperparameters that balance each loss item.

By fully leveraging the MLP's semantic aggregation and storage capabilities, we can extend the M²IV self-distillation framework to many-shot ICL scenarios, overcoming the context window limitations of LVLM. Here, each context in $\mathcal{D}_\mathbf{C}$ contains more than 100 demonstrations, which are partitioned into overlapping windows of length $w$ with overlap $o$. Each subcontext is processed individually by the teacher model, and the outputs of its MLP branches are extracted as the semantic representation for that window. These representations are then sequentially aggregated in a pairwise manner until a final, comprehensive representation is obtained that encapsulates the semantic information of the entire $n$-shot context and serves as the teacher model's final MLP state. The detailed processing pipeline for many-shot M²IV is presented in Appendix \ref{many-shot}.

\subsection{VLibrary: M²IV Storage}
\label{3.3}
Using the above training strategy, we obtain a $\Theta^{\mathcal{D}}_{\mathcal{M}}$ for each dataset $\mathcal{D}$ and LVLM $\mathcal{M}$. To facilitate management and fully exploit M²IV's plug-and-play potential, we build \textbf{VLibrary}—a repository for storing learned vectors after training. Each M²IV is indexed by its $\boldsymbol{\alpha}^a$. When wishing to equip an LVLM with domain-specific knowledge or steer it toward a desired generation pattern, we can retrieve the corresponding M²IV from VLibrary by its index and inject it into the model according to Eq~\ref{injection}. VLibrary is designed solely for storing and indexing the trained vectors and performs no inter-vector operations. As a result, the structural requirements of VLibrary remain minimal and consistent across LVLMs with different hidden state dimensions. We can implement VLibrary at minimal cost, and its practical details are provided in Appendix \ref{vli}. By translating the gains of M²IV into practical utility, VLibrary constitutes an integral part of the overall M²IV framework.
\section{Experiment}
\label{exp}
\subsection{Implementation Details}
\textbf{Benchmarks.} For VQA, we select three widely used datasets: VQAv2 \citep{vqav2}, VizWiz \citep{vizwiz}, and OK-VQA \citep{okvqa}. Towards more complex VL scenarios, we incorporate A-OKVQA \citep{A-okvqa} and GQA \citep{gqa}, which emphasize multi-hop reasoning \citep{multi-hop}. Additionally, we include the Asia split of the multicultural VQA benchmark CVQA \citep{cvqa} to assess the ability to integrate novel in-domain knowledge into pretrained models. For an evaluation of the general ICL, we utilize the image-to-text split of the latest multimodal ICL benchmark, VL-ICL bench \citep{vlicl}. To better serve as inputs for ICL, we make necessary modifications to these benchmarks; details are provided in Appendix \ref{app:dataset}.

\textbf{Configurations.} We evaluate three LVLMs: OpenFlamingov2 (9B), Idefics2 (8B), and LLaVA-NeXT (7B) \citep{llava}. These models differ in their LLM backbones, connection modules, and context windows\footnote{We provide results on three more recent LVLMs, LLaVA-OneVision (7B), InternVL2.5 (8B), and Qwen2.5VL (7B), in Appendix \ref{additionalexp}.}. Unless otherwise noted, all results are reported as the \textbf{average} across these models. Following §\ref{3.2}, we construct a query set $\mathcal{D}_{\mathbf{Q}}$ (of size $K$) and a context set $\mathcal{D}_\mathbf{C}$ (of size $2K$) from each benchmark’s training set, with $K$ varying by benchmark. We adopt \textbf{Random Sampling} as the retrieval strategy $\mathcal{R}$, and fix the number of shots \footnote{16-shot has been empirically shown to approximate optimal performance among the chosen LVLMs \citep{16}. It also allows higher-resolution images without exceeding context windows.} $n$ to 16. During training, we use AdamW as the optimizer. Evaluation is performed on the corresponding validation sets. Initialization of $\mathbf{V}$ and $\boldsymbol{\alpha}$, along with the data sizes and hyperparameters for each benchmark, are detailed in Appendix \ref{app:training}.

\textbf{Comparative Methods.} In addition to the zero-shot baseline and $n$-shot Vanilla ICL, we compare M²IV with the representation engineering methods for ICL introduced in §\ref{section2}, including TV, FV, ICV and I2RL. All of these methods are training-free. We highlight LIVE as the key comparison in our experiments, since it also employs a training strategy to obtain layer-wise vectors. Full details of all baselines are provided in Appendix \ref{app:baseline}.

\begin{table}[t]
\vspace{-20pt}
\centering
\resizebox{\textwidth}{!}{%
\begin{tabular}{ c c c c c c c c}
\toprule
\textbf{Methods} & \textbf{VQAv2} & \textbf{VizWiz} & \textbf{OK-VQA} & \textbf{GQA} & \textbf{A-OKVQA} & \textbf{CVQA} & \textbf{VL-ICL bench} \\
\midrule
Zero-shot & 43.10 & 20.77 & 31.04 & 52.42 & 32.97 & 31.09 & 11.39 \\
Vanilla ICL & 60.08 & \underline{39.88} & 51.37 & \underline{69.70} & 50.92 & \underline{56.39} & 29.08 \\
TV & 47.88 & 23.86 & 38.24 & 60.26 & 37.58 & 43.02 & 18.42 \\
FV & 47.09 & 23.40 & 38.96 & 59.33 & 38.43 & 41.82 & 18.32 \\
ICV & 44.71 & 20.83 & 35.98 & 54.80 & 32.58 & 38.45 & 11.92 \\
I2CL & 52.02 & 26.16 & 44.13 & 61.93 & 39.10 & 49.42 & 20.48 \\
LIVE & \underline{62.22} & 39.17 & \underline{53.47} & 69.20 & \underline{51.60} & 55.42 & \underline{29.95} \\
\rowcolor[HTML]{F0F0F0}
M²IV & \textbf{63.93} & \textbf{43.40} & \textbf{55.37} & \textbf{73.81} & \textbf{53.59} & \textbf{60.11} &  \textbf{33.36}\\
\midrule
Multi-turn ICL (128-shot) & 59.60 & 38.96 & 52.61 & 66.76 & 51.25 & 58.32 & - \\
\rowcolor[HTML]{F0F0F0}
M²IV (128-shot) & 65.19 & 45.13 & 56.58 & 73.58 & 53.36 & 62.67 & - \\
Multi-turn ICL (256-shot)& 59.73 & 38.76 & 52.57 & 67.14 &51.32  & 59.59 & - \\
\rowcolor[HTML]{F0F0F0}
M²IV (256-shot) & 65.92 & 46.33 & 57.43 & 74.98 & 54.77 & 62.20 & -\\
\bottomrule
\end{tabular}%
}
\caption{\label{main}
Comparison between M²IV and baseline methods. The highest scores are highlighted in \textbf{bold} and the second-best scores are \underline{underlined}. In addition to the primary 16-shot results, this table also presents the outcomes of many-shot ICL achieved with M²IV. Detailed results for each LVLM can be found in Table \ref{detailed2}.
}
\end{table}

\subsection{Main Results}
As shown in Table \ref{main}, training-free methods outperform the zero-shot baseline but still fall short of 16-shot Vanilla ICL, indicating limited capacity for handling complex multimodal tasks. While LIVE’s training strategy yields improved performance, it still underperforms on benchmarks such as VizWiz, GQA, and CVQA. In contrast, M²IV achieves the \textbf{best} scores on all benchmarks, surpassing 16-shot Vanilla ICL by an average of 3.74\%, and outperforming LIVE by 3.52\%, 4.11\%, and 3.72\% on the benchmarks where LIVE lags behind. As shown in Table \ref{detailed2}, our method ranks \textbf{first} in 18 out of 21 experiments across all three LVLMs. These results confirm that M²IV effectively preserves semantic fidelity and further distills fine-grained information, thereby outperforming Vanilla ICL and other comparative methods in a variety of complex multimodal ICL tasks. Paired with VLibrary, our method enables efficient and precise LVLM steering. Moreover, M²IV requires only about 24\% of the data used by LIVE, underscoring its efficiency and potential in data-limited scenarios. In many-shot scenarios, M²IV also yields consistent and significant improvements, demonstrating that the aggregation of MLP outputs counteracts multi-turn forgetting and effectively unlocks the benefits of many-shot ICL.

\subsection{Efficiency Analysis}
\begin{figure}[t]
    \centering
    \begin{minipage}[t]{0.48\textwidth}
        \centering
        \includegraphics[width=\linewidth]{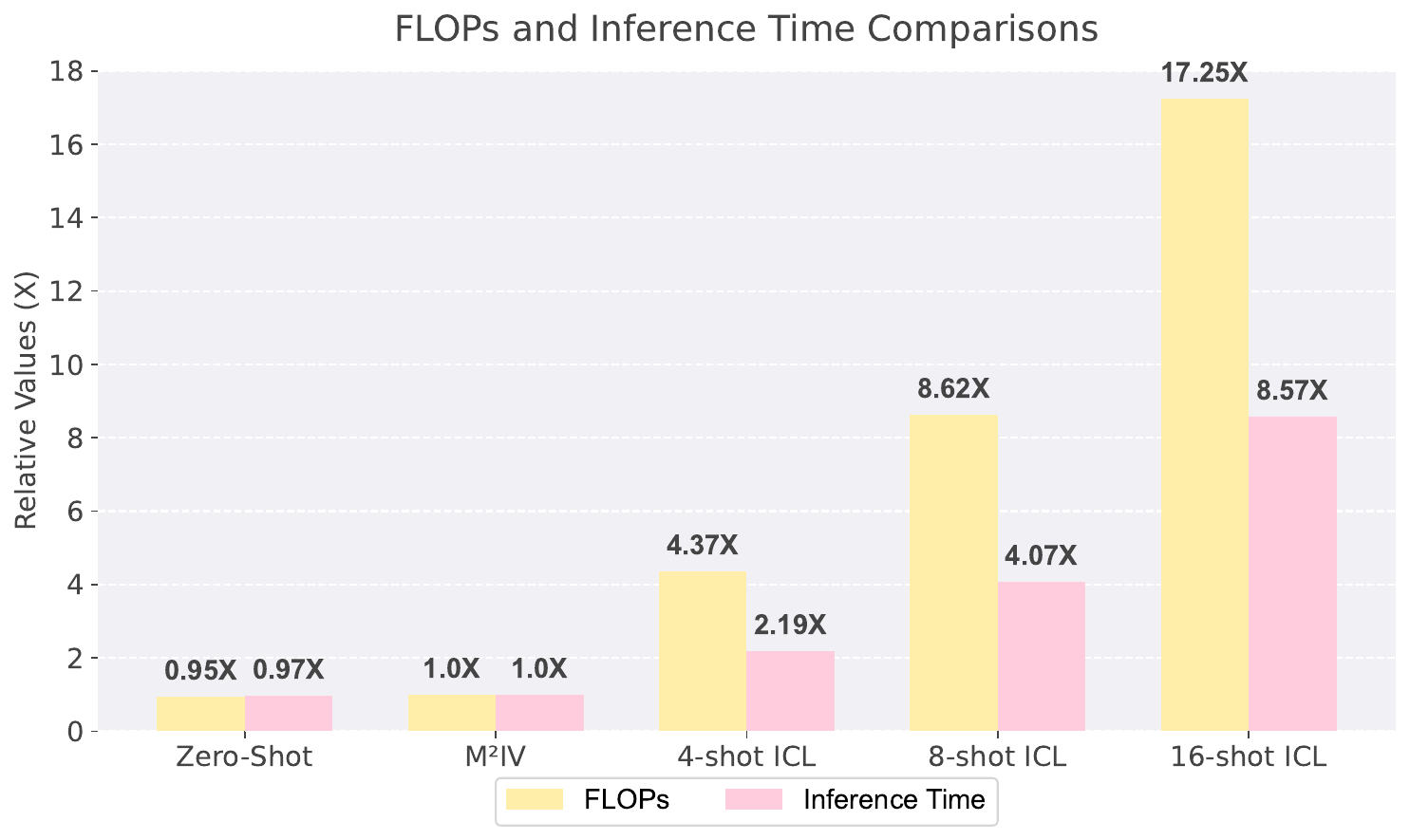}
        \caption{Comparison of M²IV, Zero-shot, and $n$-shot Vanilla ICL in terms of the total number of FLOPs and inference time used during the entire evaluation process.}
        \label{fig:effi}
    \end{minipage}
    \hfill
    \begin{minipage}[t]{0.48\textwidth}
        \centering
        \includegraphics[width=\linewidth]{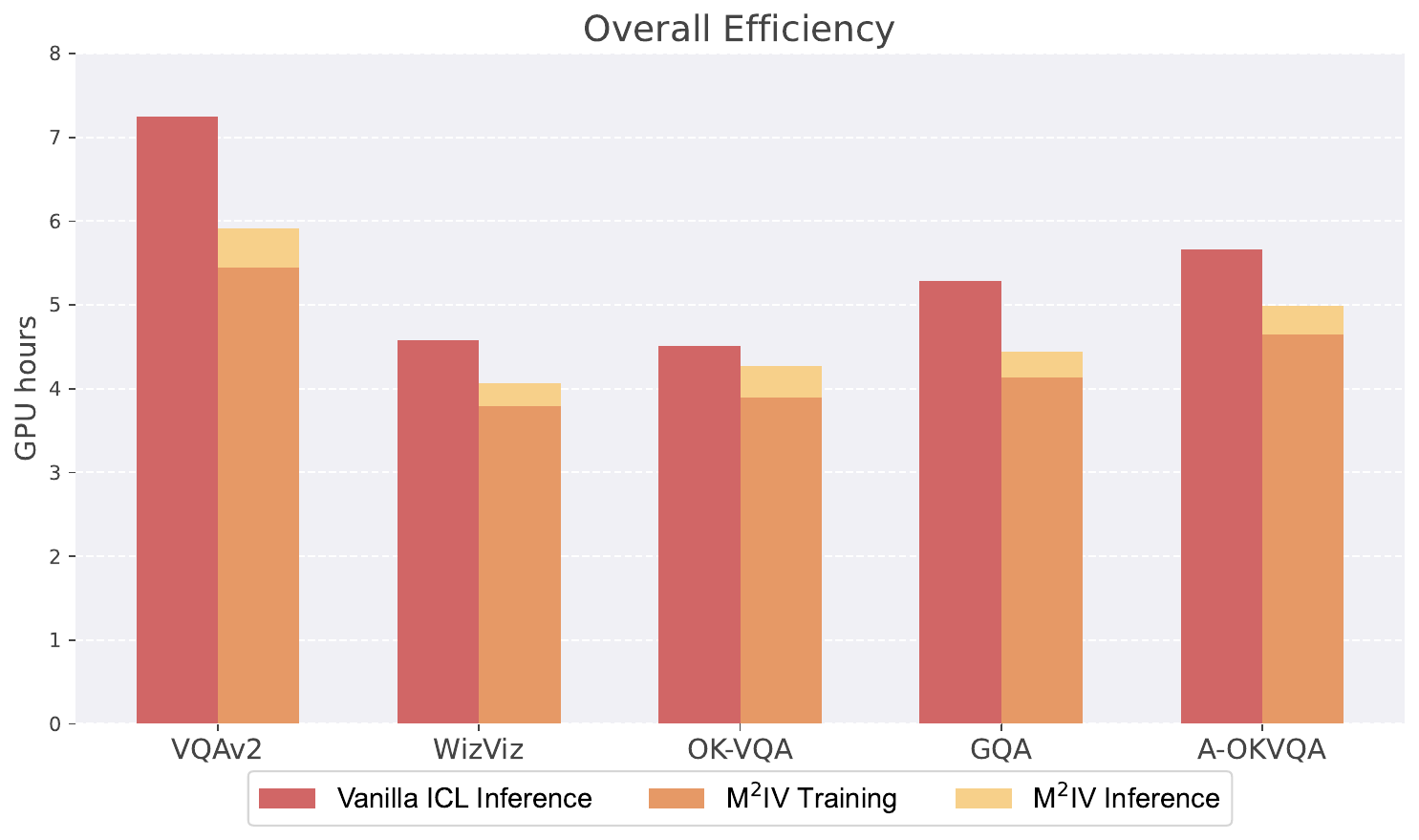}
        \caption{Comparison of GPU hours consumed by Vanilla ICL for inference across the entire evaluation and by M²IV for both training and inference.}
        \label{fig:GPU}
    \end{minipage}
\end{figure}
A principal reason for the wide adoption and growing importance of ICL is its efficiency without any parameter update. Thus, we must examine whether M²IV undermines this key advantage. First, in Figure \ref{fig:effi} we observe that M2IV preserves and even improves the inference efficiency of ICL thanks to its plug-and-play design. Its direct internal injection eliminates the large token overhead introduced by explicit in-prompt demonstrations, thus cutting FLOPs and inference time to levels well below Vanilla ICL and approaching those seen in zero-shot settings. Beyond inference latency, we also report the overall runtime that includes the training cost of M²IV. This metric can better reflect the true efficiency of the competing methods. As shown in Figure \ref{fig:GPU}, M²IV’s inference-time savings fully offset its training cost. Moreover, as M²IV is increasingly applied to zero-shot inference tasks, its cost-effectiveness improves, making it well-suited for mass applications. It is worth noting that when the repeated demonstration-retrieval cost of Vanilla ICL is taken into account, the advantage of M²IV becomes even more significant, as each knowledge base needs only a single end-to-end training run to produce a reusable $\Theta$ for the target LVLM. These results collectively demonstrate that, although M²IV involves a training phase, it delivers \textbf{higher efficiency} than Vanilla ICL. In Appendix \ref{app:method}, we compare M²IV with LoRA and prefix tuning, highlighting its joint benefits in precision and efficiency.

\subsection{Ablation Studies and Discussions} 
\label{main:ablation}
In this section, we discuss three primary concerns through extensive ablation experiments. 

\begin{figure}[t]
\vspace{-20pt}
    \centering
    \begin{minipage}[t]{0.48\textwidth}
        \vspace{0pt}
        \centering
        \includegraphics[width=\linewidth]{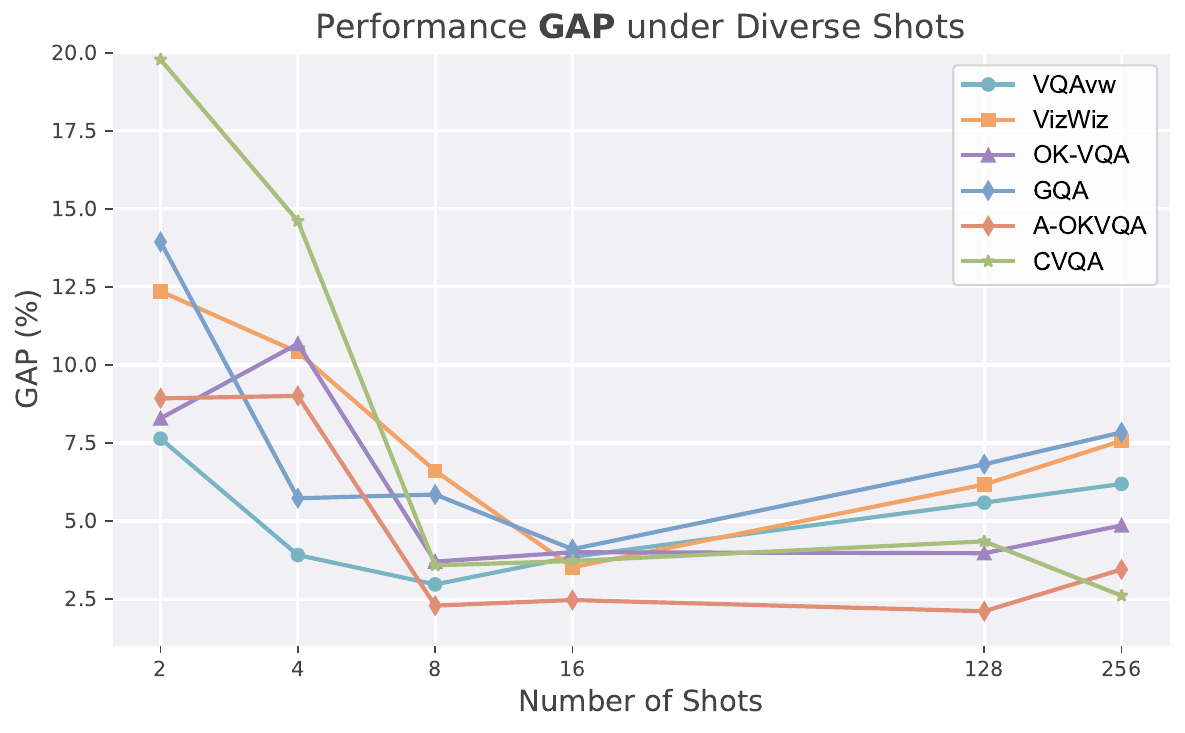}
        \caption{Performance gap over same shot Vanilla ICL (few-shot) or multi-turn ICL (many-shot) across different shot settings. Each value indicates how much M²IV outperforms the baseline.}
        \label{fig-shot}
    \end{minipage}
    \hfill
    \begin{minipage}[t]{0.48\textwidth}
        \vspace{0pt}
        \centering
        \includegraphics[width=\linewidth]{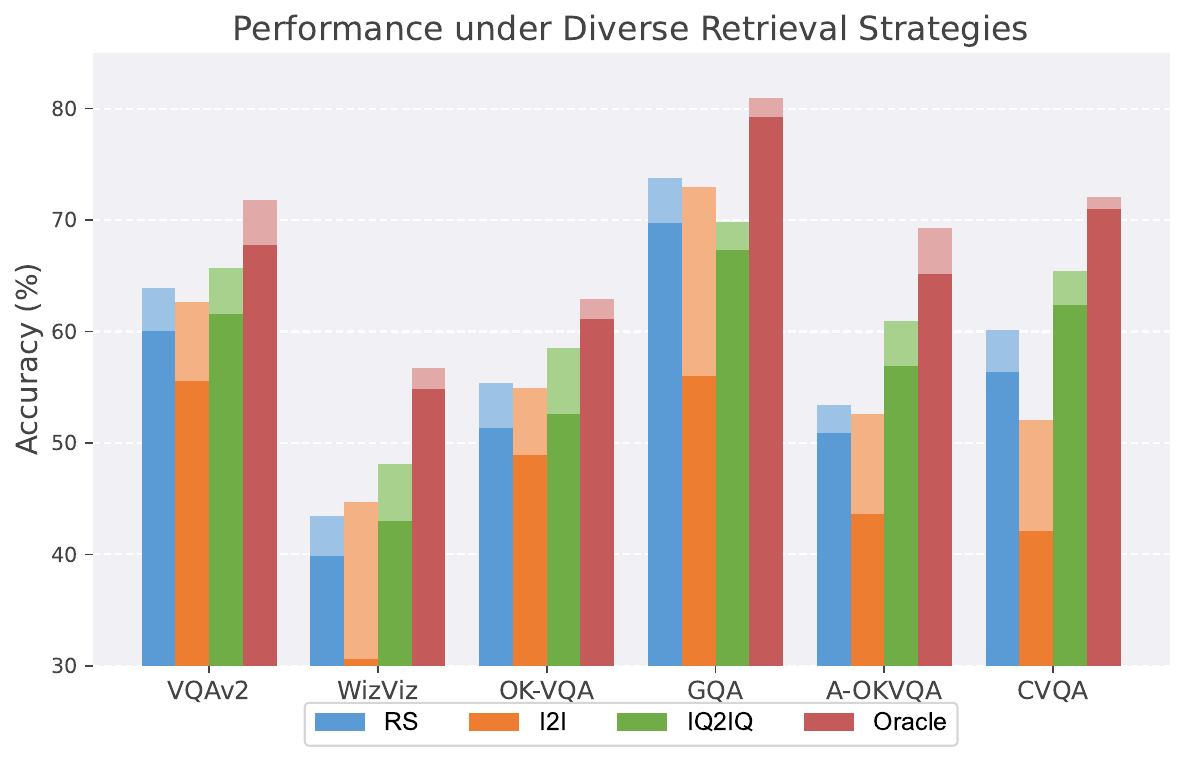}
        \caption{Performance of Vanilla ICL and M²IV across diverse demonstration retrieval strategies. The light-colored portion of each bar indicates the gain achieved by M²IV.}
        \label{fig-method}
    \end{minipage}
    \vspace{-5pt}
\end{figure}

\textbf{Why does M²IV not only replicate the effect of Vanilla ICL but also surpass it?} We first explore the advantages of M²IV over Vanilla ICL as the shot count varies. As shown in Figure \ref{fig-shot}, M²IV consistently surpasses Vanilla ICL across all shot settings, highlighting the robustness of its fine-grained representation. Notably, the largest gains occur in 2-shot and 4-shot settings, especially on datasets with imbalanced training distributions such as CVQA and VizWiz. This suggests that M²IV can, through training, effectively capture and internalize the overall distribution of a dataset, thereby avoiding the influence of skewed distributions resulting from data scarcity. In many-shot scenarios, the increase in performance gains with higher shot counts is due to M²IV's capability of mitigating the forgetting issue that worsens with additional input turns.

We further explore how the effectiveness of M²IV varies with diverse retrieval strategies  $\mathcal{R}$. Including Random Sampling (RS), we compare four strategies: I2I (image similarity-based retrieval), IQ2IQ (image-question joint similarity-based retrieval) and Oracle (a greedy retrieval performed by the LVLM based on ground-truth answers\footnote{Due to its reliance on ground-truth answer, Oracle is used to simulate optimal contexts and inapplicable in real-world scenarios. Details of these strategies are provided in Appendix \ref{app:retrieve}.}). As shown in Figure \ref{fig-method}, Vanilla ICL with I2I performs the worst across all benchmarks, echoing previous findings that I2I tends to lead to visual shortcut learning and hallucinations in complex tasks \citep{li2024configure}. Remarkably, M²IV always delivers the largest gains under this strategy, demonstrating its ability to filter out isolated or misleading features while preserving holistic semantics.

\textbf{Which component contributes most to V’s improved ICL performance?} We first investigate the data aspect in Appendix \ref{app:livecon} and find that the distribution of training data is far more influential than its sheer volume. Here, we conduct more comprehensive ablation studies on the key designs of M²IV. Table \ref{table:ablation} reveals the following points: (1) Joint embedding-based clustering improves generalization due to varied modality biases across datasets. (2) Clustering optimizes the distribution of training data while augmenting $\mathcal{D}_{\mathbf{C}}$ improves the robustness of M²IV. (3) Among training losses, $\mathcal{L}_{sup}$ yields only an average gain of 2.15\%, while $\mathcal{L}_{syn}$ delivers an average gain of 13.00\% and also outweighs all data processing strategies. Clearly, $\mathcal{L}_{syn}$ underpins our method, underscoring the importance of synergizing the MHA and MLP branches to ensure general semantic fidelity and fine-grained filtering.

\begin{table}[t]
\centering
\resizebox{\textwidth}{!}{%
\begin{tabular}{ c c c c c c c c}
\toprule
\textbf{} & \textbf{VQAv2} & \textbf{VizWiz} & \textbf{OK-VQA} & \textbf{GQA} & \textbf{A-OKVQA} & \textbf{CVQA} & \textbf{VL-ICL bench} \\
\midrule
\rowcolor[HTML]{F0F0F0}
M²IV & 63.93 & 43.40 & 55.37 & 73.81 & 53.59 & 60.11& 33.36 \\
\midrule
(a) $I$ only & 50.48& 32.61& 52.07& 68.26& 47.41 & 43.06& 32.80\\
(b) $Q$ only & 58.72& 38.97& 53.29& 70.62& 48.94 &59.25  & 32.72\\
(c) w/o clustering & 51.98& 29.74 & 47.58 & 66.37& 45.36 & 44.05& 30.31\\
(d) w/o augmentation & 55.48 & 33.28 & 51.13 & 69.61& 43.23 & 54.81 & 30.59 \\
\midrule
(e) w/o $\mathcal{L}_{mim}$ & 56.87 & 32.42 & 45.79 & 64.95 & 43.91 & 51.33 & 24.39 \\
(f) w/o $\mathcal{L}_{syn}$ & 53.13 & 28.61 & 47.30 & 58.92 & 38.79 & 43.17 & 22.63 \\
(g) w/o $\mathcal{L}_{sup}$& 61.53 & 39.64 & 53.90 & 72.02 & 50.79 & 59.83 & 30.81\\
\bottomrule
\end{tabular}%
}
\caption{\label{table:ablation}
Accuracy (\%) of M²IV under diverse ablation settings. (a)–(d) focus on the data processing phase: (a) and (b) use only image or question similarity for clustering; (c) replaces the $K$-means clustering with random sampling; (d) omits shuffled augmentation of $\mathcal{D}_\mathbf{C}$. (e)–(g) each remove one of the training loss terms.
}
\vspace{-10pt}
\end{table}

\textbf{M²IV seems to be task-specific and model-specific. Does this impact its flexibility?} The mathematical properties of vectors allow M²IV to support flexible combination and transfer, with proof provided in Appendix \ref{appproof:com}. We propose two strategies for combining multiple M²IVs to endow the model with multi-task capabilities: (1) a training-free linear addition, and (2) the introduction of a learnable parameter $\alpha$ that is fine-tuned using a small amount of multi-task data. Thus, by creating a set of atomic M²IVs, we can customize combinations to meet diverse needs. Similar strategies can also be applied to facilitate cross-model transfer of M²IV. If two LVLMs have the same number of layers and an identical hidden state dimension, an M²IV trained on one model can be inserted into the other without any additional training and will deliver comparable results. Adding a learnable parameter for light fine-tuning can further enhance this effect. The detailed procedures and results of the above strategies are collectively presented in Appendix \ref{app:cf}. They highlight the unique advantage of steering LVLMs' intermediate representations via M²IV.

\section{VLibrary: An All-purpose Toolbox for LVLM}
In this section, we demonstrate the practical value of VLibrary in solving the key challenges that LVLM faces in real-world applications. As shown in Figure \ref{fig:appli}, VLibrary empowers us to store and retrieve tailored M²IV, facilitating versatile steering while seamlessly integrating into existing systems. For the subsequent applications, we set $K=1250$ during training.

\begin{wraptable}{r}{0.5\textwidth}
\centering
\small 
\resizebox{\linewidth}{!}{
\begin{tabular}{c c c c c}
\toprule
\textbf{Methods} & \textbf{VQAv2} & \textbf{VizWiz} & \textbf{OK-VQA} & \textbf{CVQA} \\
\midrule
Zero-shot & 43.10 & 20.77 & 31.04 & 31.09 \\
\midrule
+M²IV & & & & \\
First 10 layers & \textbf{46.92}& \textbf{24.20} & \textbf{34.51} & \textbf{36.27} \\
Middle 10 layers & 43.29 & 22.09 & 32.13 & 32.41 \\
Last 10 layers & 43.82 & 21.14 & 30.19 & 30.79 \\
All layers & 45.17 & 22.05 & 32.83 & 32.86 \\
\bottomrule
\end{tabular}
}
\caption{Performance on four VQA benchmarks in the zero-shot settings and with M²IV for captioning injected at various positions.}
\vspace{-10pt}
\label{tab:layer}
\end{wraptable}
\textbf{VLibrary can enhance cross-modal alignment in a special way.} In §\ref{3.1}, we configure M²IV on a layer-wise basis, recognizing that each layer contributes differently to ICL. Alternatively, M²IV can be injected only into layers chosen for their functional importance; for instance, targeting shallow layers to reinforce fine-grained cross-modal alignment. To evaluate this, we construct detail-focused datasets by prompting GPT-4o to expand MSCOCO's \citep{coco} captions into detailed descriptions of each image’s visual features, and train M²IV on them. We then inject part of the trained M²IV into corresponding LVLM layers for zero-shot VQA. As shown in Table \ref{tab:layer}, injecting M²IV into the first 10 layers yields the greatest performance gains, even surpassing full injection. These findings reveal that M²IVs can be flexibly applied to enhance LVLMs' overall capabilities.

\begin{wraptable}{r}{0.5\textwidth}
\centering
\resizebox{\linewidth}{!}{
\begin{tabular}{c c c c c}
\toprule
\textbf{Method} & \textbf{Conversation} & \textbf{Detail} & \textbf{Complex} & \textbf{All} \\
\midrule
16-shot Vanilla ICL & 63.84& 56.41& 75.23& 67.20\\
LoRA & 69.24& 63.27& 83.05& 74.09\\
16-shot M²IV & 72.53& 64.95& 86.26 & 76.93\\
128-shot M²IV &74.34 &66.70 &84.09 & 76.89
 \\
\bottomrule
\end{tabular}
}
\caption{Instruction following evaluation of LVLM. We employ the same metrics as the original LLaVA-Bench, using GPT-4 to score the generated content.}
\vspace{-10pt}
\label{tab:if}
\end{wraptable}
\textbf{VLibrary enables versatile customization of LVLM outputs.} \textit{Instruction following} is vital for LVLMs to align with user intent and facilitate various interactions. However, the added visual modality complicates instruction adherence, requiring extensive parameter updates to achieve proper alignment. We use LLaVA-Bench to test whether M²IV can steer the model to follow specific user instructions when generating content. The benchmark covers three types of instruction: conversation, detailed description, and complex reasoning \citep{Llava1}. We train M²IV on the LLaVA dataset, which contains the same three instruction types. As shown in Table~\ref{tab:if}, M²IV consistently enhances instruction-following performance across all types, emphasizing its effectiveness in addressing the challenge with minimal overhead. In Appendix \ref{app:lib}, we also explore using M²IV to enable LVLMs to explicitly output their reasoning process.

\textbf{VLibrary is well-suited for studying LVLM safety.} M²IV’s strong behavior-steering capability makes it a powerful tool for investigating jailbreak scenarios with LVLMs. By constructing M²IV vectors that encode harmful multimodal instructions and injecting them into a model, we can compel it to override its moral safeguards and generate disallowed content. In the opposite direction, safety-oriented M²IV vectors strengthen LVLM’s awareness of harmful prompts, allowing it to detect and refuse such requests with greater precision. Experimental results and detailed analyses appear in Appendix \ref{app:lib}.
\vspace{-10pt}
\section{Conclusion}
In this study, we present M²IV, a novel representation engineering method for multimodal ICL. It leverages the unique roles of MHA and MLP branches in residual streams. Through training, M²IV achieves complex multimodal understanding and fine-grained semantic distillation, demonstrating SOTA performance on three LVLMs and seven benchmarks with relatively limited training data while maintaining ICL's efficiency. The retrieval-then-injection design of VLibrary further expands M²IV's applicability, enabling rapid solutions to many practical challenges in LVLM. In general, M²IV offers a promising paradigm for both multimodal ICL and LVLM steering, providing valuable insights for further breakthroughs in the multimodal domain. Currently, our method is only applicable to open-source LVLMs. In future work, we hope to extend to closed-source LVLMs, possibly by utilizing the trained M²IVs to empower a lightweight language model dedicated to demonstration selection.
\vspace{-10pt}
\section*{Acknowledgments}
We sincerely thank Tian Yun and Ellie Pavlick from Brown University for their helpful suggestions on this paper.

\section*{Ethics Statement}
This paper presents examples that contain harmful or offensive language and imagery, primarily drawn from HatefulMemes and MM-SafetyBench. We unequivocally reject the disrespectful and unethical views expressed in these materials, which appear here only to advance the creation of a fair, unbiased, and safe community for large vision-language models.

\bibliography{colm2025_conference}
\bibliographystyle{colm2025_conference}

\appendix

\section{Additional Theoretical Proof}
\subsection{Proof of Theorem \ref{mha}}
\label{proof_mha}
\begin{proof} 
The attention mechanism for query vector $\mathbf{h}^{i}$ over a key-value pair $\begin{bmatrix} \mathbf{C}^{\top} \ \mathbf{Q}^{\top} \end{bmatrix}^{\top}$ is:
\begin{equation}
    \text{Attn}\left(\mathbf{h}^{i}, \begin{bmatrix}
    \mathbf{C}^{\top}
    \:
    \mathbf{Q}^{\top}
    \end{bmatrix}^{\top}, \begin{bmatrix}
    \mathbf{C}^{\top}
    \:
    \mathbf{Q}^{\top}
    \end{bmatrix}^{\top}\right)
    =\text{softmax}\left(\begin{bmatrix}
     d_{\mathcal M}^{-\frac{1}{2}}{\mathbf{h}^{i}\mathbf{C}^{\top }} & 
     d_{\mathcal M}^{-\frac{1}{2}}{\mathbf{h}^{i}\mathbf{Q}^{\top }}
    \end{bmatrix}\right)\begin{bmatrix}\mathbf{C}\\\mathbf{Q}\end{bmatrix},
\end{equation}
where $\text{softmax}\left(\cdot\right)$ normalizes the input into a probability distribution, and $d_{\mathcal M}^{-\frac{1}{2}}$ is the scaling factor to stabilize training. Expanding the computation, let: 
\begin{equation}
s_{\mathbf{C}} = \sum_{j=1}^{C} \text{exp}\left(d_{\mathcal M}^{-\frac{1}{2}}{\mathbf{h}^{i} \mathbf{C}_{j}^{\top}}\right),\quad
s_{\mathbf{Q}} = \sum_{j=1}^{I} \text{exp}\left(d_{\mathcal M}^{-\frac{1}{2}}{\mathbf{h}^{i} \mathbf{Q}_{j}^{\top}}\right),
\end{equation}
which denotes the sums of exponentiated scores over the demonstration and query tokens, respectively, where $\mathbf{C}_{j}$ and $\mathbf{Q}_{j}$ are the $j$-th rows of $\mathbf{C}$ and $\mathbf{Q}$. Thus, the attention output is:
\begin{align}
&\text{Attn}\left(\mathbf{h}^{i}, \begin{bmatrix}
    \mathbf{C}^{\top}
    \:
    \mathbf{Q}^{\top}
    \end{bmatrix}^{\top}, \begin{bmatrix}
    \mathbf{C}^{\top}
    \:
    \mathbf{Q}^{\top}
    \end{bmatrix}^{\top}\right) \nonumber \\
&= \left(s_{\mathbf{C}} + s_{\mathbf{Q}}\right)^{-1} \left( \text{exp}\left(d_{\mathcal{M}}^{-\frac{1}{2}}{\mathbf{h}^{i} \mathbf{C}^{\top}}\right) \mathbf{C} + \text{exp}\left(d_{\mathcal{M}}^{-\frac{1}{2}}{\mathbf{h}^{i} \mathbf{Q}^{\top}}\right) \mathbf{Q} \right) \nonumber \\
&= {s_{\mathbf{C}}}\left(s_{\mathbf{C}} + s_{\mathbf{Q}}\right)^{-1} \text{softmax}\left(d_{\mathcal{M}}^{-\frac{1}{2}}{\mathbf{h}^{i} \mathbf{C}^{\top}}\right) \mathbf{C}
+
{s_{\mathbf{Q}}}\left(
s_{\mathbf{C}} + s_{\mathbf{Q}}\right)^{-1} \text{softmax}\left(d_{\mathcal{M}}^{-\frac{1}{2}}{\mathbf{h}^{i} \mathbf{Q}^{\top}}\right) \mathbf{Q}.
\end{align}
Based on the aforementioned analysis, we take $\Psi$, $\zeta^i$, and $\eta^i$ as follows:
\begin{equation}
    \Psi\left(\varrho _q,\varrho _k,\varrho _v\right)\equiv \text{softmax}\left(d_{\mathcal{M}}^{-\frac{1}{2}}{\varrho _q\varrho _k^{\top }}\right)\varrho _v, 
    \quad \forall \left(\varrho _q,\varrho _k,\varrho _v\right),
\end{equation}
\begin{equation} \zeta^i:={s_{\mathbf{C}}}\left(s_{\mathbf{C}} + s_{\mathbf{Q}}\right)^{-1},\quad
\eta^i:=s_{\mathbf{Q}}\left(s_{\mathbf{C}} + s_{\mathbf{Q}}\right)^{-1},
\end{equation}
and then the proof of the theorem is completed.
\end{proof}
\subsection{Proof of Theorem \ref{mlp}}
\label{proof_mlp}
\begin{proof}
    Based on Theorem \ref{mha}, there exist $\zeta^{i}$, $\eta^i \in\mathbb{R}$ satisfying:
    \begin{equation}\label{eq:attention_output}
        \mathbf{a}^i_{\mathbf{C},\mathbf{Q}}=\zeta^i\cdot \mathbf{a}^i_{\mathbf{C}}+\eta^i\cdot \mathbf{a}^i_{\mathbf{Q}}.
    \end{equation}
    Multiply both sides of Eq \eqref{eq:attention_output} on the right by $\mathbf{W}$, we obtain the following equation:
    \begin{equation}
    \mathbf{a}^i_{\mathbf{C},\mathbf{Q}}\mathbf{W}=\zeta^i\cdot \mathbf{a}^i_{\mathbf{C}}\mathbf{W}+\eta^i\cdot \mathbf{a}^i_{\mathbf{Q}}\mathbf{W}.
    \end{equation}
    Take $\psi_{\mathbf{W}}\left(\mathbf{x}\right)\equiv \mathbf{x}\mathbf{W}$, we obtain:
    \begin{equation}
    \text{MLP}\left(\mathbf{a}^i_{\mathbf{C},\mathbf{Q}}\right)=\zeta^i \cdot \psi_{\mathbf{W}}\left(\mathbf{a}^i_{\mathbf{C}}\right) + \eta^i \cdot \text{MLP}\left(\mathbf{a}^i_{\mathbf{Q}}\right),
    \end{equation}
    and then the proof of the theorem is completed.
\end{proof}

\subsection{Extension to MLP with Non-linear Activations}
\begin{theorem}
    For linear transformation matrix $\mathbf{W}_1 \in \mathbb{R}^{d_{\mathcal{M}} \times d_{\text{ff}}}$, $\mathbf{W}_2 \in \mathbb{R}^{d_{\text{ff}}\times d_{\mathcal{M}}}$, there exists $\rho_{\mathbf{W}_1,\mathbf{W}_2}: \mathbb{R}^{d_{\mathcal{M}}}\to \mathbb{R}^{d_{\mathcal{M}}}$ such that, for any token position $i$, there exist $\zeta^{i}$, $\eta^{i}\in \mathbb{R}$ satisfying:
    \begin{equation}
    \mathrm{MLP}\left(\mathbf{a}^i_{\mathbf{C},\mathbf{Q}}\right)=\zeta^i\cdot \rho_{\mathbf{W}_1,\mathbf{W}_2}\left(\mathbf{a}^i_{\mathbf{C}}\right)+\eta^i\cdot \mathrm{MLP}\left(\mathbf{a}^i_{\mathbf{Q}}\right),
    \end{equation}
    where the MLP operation with non-linear activations is defined as $\mathrm{MLP}\left(x \right)\equiv \sigma\left(x\mathbf{W}_1\right)\mathbf{W}_2$, and $\sigma \left(\cdot \right)$ denotes the activation function.
\end{theorem}
\begin{proof}
        Based on Theorem \ref{mha}, there exist $\zeta^{i}$, $\eta^i \in\mathbb{R}$ satisfying:
    \begin{equation}\label{eq:attention_output_non-linear}
        \mathbf{a}^i_{\mathbf{C},\mathbf{Q}}=\zeta^i\cdot \mathbf{a}^i_{\mathbf{C}}+\eta^i\cdot \mathbf{a}^i_{\mathbf{Q}}.
    \end{equation}
    Multiply both sides of Eq \eqref{eq:attention_output_non-linear} on the right by $\mathbf{W}_1$, we obtain the following equation:
    \begin{equation}
    \mathbf{a}^i_{\mathbf{C},\mathbf{Q}}\mathbf{W}_1=\zeta^i\cdot \mathbf{a}^i_{\mathbf{C}}\mathbf{W}_1+\eta^i\cdot \mathbf{a}^i_{\mathbf{Q}}\mathbf{W}_1.
    \end{equation}
    Apply $\sigma \left(\cdot \right)$ to both sides of the equation and then multiply by $\mathbf{W}_2$, we obtain:
    \begin{equation}
    \sigma\left(\mathbf{a}^i_{\mathbf{C},\mathbf{Q}}\mathbf{W}_1\right)\mathbf{W}_2=\zeta^i\cdot \sigma\left(\mathbf{a}^i_{\mathbf{C}}\mathbf{W}_1\right)\mathbf{W}_2+\eta^i\cdot \sigma\left(\mathbf{a}^i_{\mathbf{Q}}\mathbf{W}_1\right)\mathbf{W}_2.
    \end{equation}
    Take $\rho_{\mathbf{W}_1,\mathbf{W}_2}\left(\mathbf{x}\right)\equiv \sigma\left(\mathbf{x}\mathbf{W}_1\right)\mathbf{W}_2$, we obtain:
    \begin{equation}
    \text{MLP}\left(\mathbf{a}^i_{\mathbf{C},\mathbf{Q}}\right)=\zeta^i \cdot \rho_{\mathbf{W}_1,\mathbf{W}_2}\left(\mathbf{a}^i_{\mathbf{C}}\right) + \eta^i \cdot \text{MLP}\left(\mathbf{a}^i_{\mathbf{Q}}\right),
    \end{equation}
    and then the proof of the theorem is completed.
\end{proof}

\subsection{Task Combination}
\label{appproof:com}
Each learned in-context vector encodes the contextual semantics of different tasks. We further demonstrate that by linearly combining these vectors, we can obtain the context required for new tasks. This property of linear composability enhances the representational capacity of in-context vectors, thereby improving the generalization capability of the LVLM. Given $n\in \mathbb{N}_{+}$ as the number of tasks, a model $\mathcal M$ with hidden dimension $d_{\mathcal M}$, and matrices
$\left \{\mathbf{C}_t\right \}_{t=1}^{n}$ representing the in-context demonstrations for each task, we investigate whether these demonstrations could be combined to support multimodal ICL across multiple tasks.

\begin{theorem}\label{thm:combine}
There exists a function $\phi:\mathbf{D}_{\phi}\to \mathbb{R}^{d_{\mathcal{M}}}$ such that for any query matrix $\mathbf{Q}$ and token position $i$ corresponding to the residual stream $\mathbf{h}^{i}$, there exist $\left \{\vartheta^i_t\in \mathbb{R}\right \}_{t=1}^{n}$ and $\varpi^i\in \mathbb{R}$ such that:
\begin{align}
    & \mathrm{Attn}\left(
    \mathbf{h}^{i}
    , 
    \left [
    \mathbf C_1^{\top} \: \mathbf C_2^{\top} \: \dots \: \mathbf C_n^{\top} \: \mathbf{Q}^{\top}
    \right ]^{\top}
    , 
    \left [
    \mathbf C_1^{\top} \: \mathbf C_2^{\top} \: \dots \: \mathbf C_n^{\top} \: \mathbf{Q}^{\top}
    \right ]^{\top}
    \right)  \nonumber \\
    & =\sum_{t=1}^{n} \vartheta^i_t\cdot \phi \left(\mathbf{h}^{i}, \mathbf{C}_t, \mathbf{C}_t\right)
    +
    \varpi^i\cdot \mathrm{Attn}\left(\mathbf{h}^{i}, \mathbf{Q}, \mathbf{Q}\right).
\end{align}
\end{theorem}

\begin{proof}
Based on Theorem \ref{mha} and the computational formula of the attention mechanism, for any query matrix $\mathbf{Q}$ and any token position $i$ in the residual stream $\mathbf{h}^{i}$, we obtain:
\begin{align}
    & \text{Attn}\left(
    \mathbf{h}^{i}
    , 
    \left[
    \mathbf C_1^{\top} \: \mathbf C_2^{\top} \: \dots \: \mathbf C_n^{\top} \: \mathbf{Q}^{\top}
    \right]^{\top}
    , 
    \left[
    \mathbf C_1^{\top} \: \mathbf C_2^{\top} \: \dots \: \mathbf C_n^{\top} \: \mathbf{Q}^{\top}
    \right]^{\top}
    \right)  \nonumber \\
    & = \zeta^i_1\cdot \Psi \left(\mathbf{h}^{i}, \mathbf{C}_1, \mathbf{C}_1\right) +\eta^i_1\cdot \text{Attn}\left(
    \mathbf{h}^{i}
    , 
    \left[
    \mathbf C_2^{\top} \: \dots \: \mathbf C_n^{\top} \: \mathbf{Q}^{\top}
    \right]^{\top}
    , 
    \left[
       \mathbf C_2^{\top} \: \dots \: \mathbf C_n^{\top} \: \mathbf{Q}^{\top} 
    \right]^{\top}
    \right)
    \nonumber \\ 
    & = \zeta^i_1\cdot \Psi \left(\mathbf{h}^{i}, \mathbf{C}_1, \mathbf{C}_1\right)+\eta^i_1\zeta^i_2\cdot \Psi \left(\mathbf{h}^{i}, \mathbf{C}_2, \mathbf{C}_2\right) \nonumber \\
     & \hspace{1em} + \eta^i_1\eta^i_2\cdot \text{Attn}\left(
    \mathbf{h}^{i}
    , 
    \left [
    \mathbf C_3^{\top} \: \dots \: \mathbf C_n^{\top} \: \mathbf{Q}^{\top}
    \right ]^{\top}
    , 
    \left [
       \mathbf C_3^{\top} \: \dots \: \mathbf C_n^{\top} \: \mathbf{Q}^{\top} 
    \right]^{\top}
    \right) \nonumber \\
    & = \dots \nonumber \\
    & = \sum_{t=1}^{n} \left(\prod_{k=1}^{t} \eta^i_{k-1}\right)\zeta^i_t\cdot \Psi \left(\mathbf{h}^{i}, \mathbf{C}_t, \mathbf{C}_t\right)+\left(\prod_{t=1}^{n} \eta^i_{t}\right)\cdot \text{Attn}\left(\mathbf{h}^{i}, \mathbf{Q}, \mathbf{Q}\right),
\end{align}
where $\zeta_t^i$ and $\eta_t^i$ are derived from the attention scores for each task's demonstration matrix and the query matrix, respectively. Take $\phi$, $\left \{\vartheta^i_t\right \}_{t=1}^{n}$, and $\varpi^i$ as follows:
\begin{equation}
    \phi\left(\varrho _q,\varrho _k,\varrho _v\right)\equiv \Psi\left(\varrho _q,\varrho _k,\varrho _v\right), 
    \quad \forall \left(\varrho _q,\varrho _k,\varrho _v\right),
\end{equation}
\begin{equation}
    \vartheta^i_t:=\zeta^i_t\cdot \prod_{k=1}^{t} \eta^i_{k-1}, \quad \forall t \in \left \{ 1,2,\dots,n \right \} ,
\end{equation}
\begin{equation}
    \varpi ^i:=\prod_{t=1}^{n} \eta^i_{t},
\end{equation}and then the proof of the theorem is completed.
\end{proof}
\section{Dataset Processing}
\label{app:data}

For efficient training of the M²IV framework, we propose a data sampling strategy for multimodal datasets.

For each instance $(I_j, Q_j, A_j)$ in $\mathcal{D}$, we construct its semantic representation by:
\begin{equation}
    \mathbf{E}_j = \text{CLIP}(I_j) \oplus \text{CLIP}(Q_j), \quad \forall j\in\{1,2,\dots,|\mathcal{D}|\},
\end{equation}
where $\mathbf{E}_j$ denotes the joint embedding of the $j$-th sample, and $\oplus$ represents the concatenation operation. We use a CLIP-L/14 model to separately embed both the image $I_j$ and question $Q_j$, then concatenate these embeddings to form a multimodal representation. For datasets containing some long textual prompts, such as MM-SafetyBench in Appendix \ref{app:lib}, the text sometimes exceeds CLIP’s limit of 77 tokens. In such cases, we switch to LongCLIP \citep{longclip}, a well-trained variant that extends CLIP to longer text, and obtain embeddings with a consistent dimension.

We apply k-means clustering on the joint embeddings of all instances using cosine similarity as the distance metric, dividing $\mathcal{D}$ into $K$ clusters:
\begin{equation}
    \{\mathbf C_{1}, \mathbf C_{2}, \dots, \mathbf C_{K}\} = \text{k-means}(\{\mathbf{E}_j\}_{j=1}^{|\mathcal{D}|}; K, \text{cosine}),
\end{equation}
where $\mathbf{C}_i$ represents the $i$-th cluster, and $K$ is a hyperparameter that specifies the number of clusters. For each cluster, we identify the instance closest to its centroid:
\begin{equation}
    j_k = \argmin_{j \in \mathbf {C}_{k}} \text{cosine\_distance}(\mathbf{E}_j, \text{centroid}(\mathbf C_k)), \quad \forall k \in \{1, 2, \dots, K\},
\end{equation}
where $j_k$ is the index of the instance closest to the centroid of cluster $\mathbf{C}_k$. These $K$ instances form our query sample set $\mathcal{D}_{\mathbf{Q}} = \{(I_{k}, Q_{k})\}_{k=1}^K$ with each answer $A_{k}$ removed, as they will be used as query samples during training.

After extracting the query samples, we define the support set $\mathcal{D}_{\text{support}}$ as all remaining instances in $\mathcal{D}$:
\begin{equation}
    \mathcal{D}_{\text{supp}} = \mathcal{D} \setminus \{(I_{k}, Q_{k}, A_{k})\}_{k=1}^K,
\end{equation}
where $\setminus$ denotes the set difference operation.

For each query sample $I_{k}, Q_{k}$ in $\mathcal{D}_{\mathbf{Q}}$, we apply a retrieval strategy $\mathcal{R}$ to select $n$ relevant instance from the support set $\mathcal{D}_{\text{support}}$, forming an $n$-shot context $\mathbf{C}_k$:
\begin{equation}
    \mathbf{C}_k = \mathcal{R}(\mathcal{D}_{\text{supp}}, (I_{k}, Q_{k}), n), \quad \forall k \in \{1, 2, \dots, K\},
\end{equation}
where $\mathcal{R}$ can be implemented as various retrieval methods such as similarity-based retrieval, random sampling, or other domain-specific selection strategies. These $K$ contexts form our initial training set $\mathcal{D}_{\mathbf{C}} = \{\mathbf{C}_k\}_{k=1}^K$.
To enhance the robustness of the MLP output to context permutation, we augment $\mathcal{D}_{\mathbf{C}}$ by creating a permuted version $\mathbf{C}'_k$ of each context $\mathbf{C}_k$, where the order of the $n$ demonstrations is randomly shuffled:
\begin{equation}
    \mathbf{C}'_k = \text{shuffle}(\mathbf{C}_k), \quad \forall k \in \{1, 2, \dots, K\}.
\end{equation}
The final augmented training set contains $2K$ instances, comprising both the original and permuted contexts:
\begin{equation}
    \mathcal{D}_{\mathbf{C}} = \{\mathbf{C}_k,\mathbf{C}'_k\}_{k=1}^K,
\end{equation}
where each training instance consists of a context (either original or permuted). Each query sample in $\mathcal{D}_{\mathbf{Q}}$ corresponds to two contexts in $\mathcal{D}_{\mathbf{C}}$. When used as input, the prompt places the $n$-shot context in front, followed by the corresponding query sample.

\section{Synergistic Loss}
\label{app:loss}
For an LVLM with $L$ layers, suppose that at layer $l$, (where $l \in \{1,2,\dots,L\}$), after injecting $\Theta$, the final outputs of MHA and MLP at that layer are $\mathbf{A}_{l}$ and $\mathbf{M}_{l}$, respectively:
\begin{equation}
\mathbf{A}_{l}(\Theta)=\mathbf{a}_{l}+\alpha_{l}^{a}\cdot \mathbf{v}_{l}^{a}, 
\end{equation}
\begin{equation}
\mathbf{M}_{l}(\Theta)=\mathbf{m}_{l}+\alpha_{l}^{m}\cdot \mathbf{v}_{l}^{m}.     
\end{equation}
We apply the following normalization to obtain $\mathbf{Z}_{l}^{a}(\Theta)$ and $\mathbf{Z}_{l}^{m}(\Theta)$:
\begin{equation}
\mathbf{Z}_{l}^{a}(\Theta)=\frac{\mathbf{A}_{l}(\Theta)}{\left \| \mathbf{A}_{l}(\Theta) \right \| } ,
\end{equation}
\begin{equation}
\mathbf{Z}_{l}^{m}(\Theta)=\frac{\mathbf{M}_{l}(\Theta)}{\left \| \mathbf{M}_{l}(\Theta) \right \| }.
\end{equation}
We compute the cross-view correlation matrix $\mathbf{M}^{l}\in \mathbb{R}^{d_\mathcal{M}\times d_\mathcal{M}}$ for the two normalized final outputs:
\begin{equation}
    \mathbf{M}^{l}=(\mathbf{Z}_{l}^{a}\left(\Theta\right))^{\top}\mathbf{Z}_{l}^{m}\left(\Theta\right).
\end{equation}
We use this matrix to compute the \textbf{synergistic loss} as follows:
\begin{equation}
\mathcal{L}_{syn} =   \sum_{l=1}^L\left(\sum_{i=1}^{d_\mathcal{M}} \left(1 - \mathbf{M}_{ii}^{l}\right)^2 + \gamma \cdot \sum_{i=1}^{d_\mathcal{M}} \sum_{j\ne i} {\mathbf{M}_{ij}^{l}}^2\right),
\end{equation}
where $\mathbf{M}_{ij}^{l}$ is the element at the $i$-th row and $j$-th column of $\mathbf{M}^{l}$ and $\gamma$ is a hyperparameter.

\section{Many-shot M²IV}
\label{many-shot}
\textbf{Pairwise aggregation pipeline.}
A long $n$-shot prompt is first divided into overlapping windows of length $w$ with overlap $o$.  
The teacher LVLM runs on every window and the MLP activations are mean-pooled into fixed-length vectors $m_i \in \mathbb{R}^d$.  
These per-window summaries are then combined left-to-right in a size-preserving two-vector loop.  
At step $i$ the current aggregate $\hat{m}_{i-1}$ is concatenated with the new window vector to form
\begin{equation}
z = [\hat{m}_{i-1},\,m_i] \in \mathbb{R}^{2d}.
\end{equation}

Two learned projections $W_g,\,W_1 \in \mathbb{R}^{d\times 2d}$ generate a gate $g$ and a candidate $c$:
\begin{equation}
\begin{aligned}
g &= \sigma\!\left(W_g z\right),\\
\check{c} &= \tanh\!\left(W_1 z\right).
\end{aligned}
\end{equation}

The aggregate is updated by
\begin{equation}
\hat{m}_{i}= g \odot \check{c} + \left(1-g\right)\odot \hat{m}_{i-1},
\end{equation}
which keeps dimensionality at $d$, matching the original MLP hidden size.  
Sharing the same tiny gating block across all $N$ windows makes the procedure $\mathcal{O}(N)$ in time and adds only $\mathcal{O}\!\left(d^{2}\right)$ parameters, so the final vector $\hat{m}_{N}$ can be injected back into the LVLM without any adapter.

\textbf{Information loss mitigation.}
The element-wise gate learned above selectively retains high-salience features from earlier aggregates.  
To further curb boundary effects, we employ a lightweight tuning strategy to determine the optimal window length $w$ and overlap $o$:
\begin{equation}
w \in \{32,\,64,\,128\},
\qquad
o \in \Bigl\{0,\,\tfrac14 w,\,\tfrac12 w\Bigr\}.
\end{equation}

This coarse grid is evaluated on a held-out set using cosine similarity between the aggregated student vector and a full-context teacher vector as a cheap proxy for fidelity.  
The top few candidates then enter a successive-halving loop that progressively allocates more budget, and the survivor is chosen by end-task accuracy adjusted for latency.
 
\section{VLibrary}
\label{vli}
In practice, VLibrary is implemented using an off-the-shelf object store, which is easily integrated into existing systems. During retrieval, the system queries using the M²IV’s designated parameter set, $\boldsymbol{\alpha}^a$. This architecture avoids the overhead of rebuilding datastores and instead supports straightforward, scalable deployment. 

VLibrary is hosted on Amazon S3, where each M²IV asset is a structured binary object that contains the learned vectors $(\mathbf{V}^a,\mathbf{V}^m)$ and the associated scalars $(\boldsymbol{\alpha}^a,\boldsymbol{\alpha}^m)$ for each decoder layer. We serialize each asset using Protocol Buffers to preserve floating point precision, and then compress it with Zstandard before uploading. To guarantee uniqueness and support lifecycle management, we use content-based addressing. Specifically, we normalize the index parameters $\boldsymbol{\alpha}^a$ by enforcing a fixed precision and deterministic layer ordering. We then compute a SHA-256 hash, referred to as the \textit{M2IV\_Content\_Hash}, which serves as the S3 object key. For efficient access, a Redis-based Mapping Service maintains associations between human-readable versioned identifiers (e.g., \textit{model\_name@v1.2:task\_name@v1.0}) and their corresponding \textit{M2IV\_Content\_Hash}. When an application requests a specific M²IV, it provides the versioned keys, retrieves the hash from Redis, and uses it to fetch the binary object from S3. The object is then decompressed and deserialized back into a usable in-memory structure. We further accelerate frequent queries via application-level caching.
\section{Benchmarks}
\label{app:dataset}
The amount of data used in our experiments is shown in Table \ref{apptab:detail}. For few-shot VQA evaluation, we adopt the following datasets/benchmarks\footnote{For datasets with multiple human-annotated labels per sample, one of them is randomly chosen as the ground-truth label in demonstrations}:
\begin{itemize} 
\item \textbf{VQAv2} is based on images from the MSCOCO dataset and features traditional question-answer pairs, assessing a model’s ability to accurately interpret both the image and the language of the question. 
\item \textbf{VizWiz} introduces additional real-world complexities with lower-quality images, questions that often lack sufficient context, and a significant portion of unanswerable queries. Models must contend with incomplete visual information and learn to handle uncertainty. 
\item \textbf{OK-VQA} focuses on questions that require external knowledge beyond the image content, serving as a benchmark for evaluating whether models can incorporate outside information to arrive at correct answers. 
\item \textbf{GQA} is a large-scale dataset focusing on compositional question answering over real-world images, serving as a benchmark for evaluating how models parse intricate scene relationships and perform multi-step reasoning.
\item \textbf{A-OKVQA} expands upon OK-VQA by offering a wider variety of question types and more demanding knowledge requirements. Notably, 30.97\% of its samples involve at least two inference hops, highlighting the dataset’s emphasis on multi-step reasoning and deeper knowledge integration. A-OKVQA samples come in two forms, multi-choice and direct answer. We opt for the latter for open-ended evaluation. A-OKVQA provides a reasoning rationale for each instance. In the main experiments, we remove the rationale from each instance, whereas in the explainability experiments, we utilize these rationales.
\item \textbf{CVQA} is a cultural-diverse, multilingual visual question-answering benchmark that gathers images and question–answer pairs from 30 countries and 31 languages, aiming to assess a model’s ability to handle both visual input and text prompts in a truly global context. It employs a multiple-choice format—one correct answer with three distractors. CVQA is divided by continent to highlight regional and linguistic diversity. We only use its Asia split, which comprises 19 types of Asian country–language pairs, such as China–Chinese, India–Hindi and Japan–Japanese. We mix the 19 types in proportion rather than computing an average across single ones.
\end{itemize}
Besides, we use the latest multimodal ICL benchmark, \textbf{VL-ICL bench} for general ICL evaluation. VL-ICL Bench is a comprehensive evaluation suite tailored for multimodal ICL, encompassing both image-to-text and text-to-image tasks. It tests a broad range of capabilities, from fine-grained perception and reasoning to fast concept binding, all using a few demonstrations. VL-ICL Bench shows meaningful improvements with more shots and highlights fundamental model limitations. In our study, we only employ the image-to-text split of VL-ICL Bench, which includes Fast Open MiniImageNet, CLEVR Count Induction, Operator Induction, TextOCR, Interleaved Operator Induction, and Matching MiniImageNet; we test each task individually and then average the performances to obtain the final score. For splits containing only images and answers, we assign a uniform question to each instance. For instance, in the Fast Open MiniImageNet split, we use “What’s in the image?” as the question.

For the VQA datasets and VL-ICL bench, we use Accuracy(\%) as the metric to assess the models' ability to provide correct answers:
\begin{equation}
Acc_{a_{i}}=\text{max}\left(1,\frac{3 \cdot  {\textstyle \sum_{k\in \left[0,9\right]}^{}}\text{match}\left(a_{i},g_{k}\right) }{10} \right),
\end{equation}
where $a_{i}$ denotes the model's generated answer, $g_{k}$ denotes the $k$-th ground true answer. $\text{match}\left(\cdot,\cdot\right)$ decides whether two answers match, if they match, the result is 1, otherwise 0.

\begin{table}[htbp]
\centering
\begin{tabular}{c c c c c}
\toprule
\textbf{Datasets} & \textbf{Training} & \textbf{Evaluation}\\
\midrule
VQAv2& 10,000 & 10,000 \\
VizWiz& 8,000 & 4,000\\
OK-VQA & 8,000 & 5,000 \\
GQA & 10,000 &  5,000 \\
A-OKVQA & 8,000 & 5,000\\
CVQA & 3,000 & 1,500\\
VL-ICL bench & 8,360 & 1,120\\
\bottomrule
\end{tabular}
\caption{\label{LIVEsize} Overview of the size distribution across the benchmarks used in LIVE.}
\end{table}

\section{Models}
\label{app:models}
In our study, we select three LVLMs that support multi-image inputs and have demonstrated multimodal ICL capabilities: OpenFlamingov2-9B, Idefics2-8B and LLaVA-NeXT-7B. Their respective configurations are shown in Table \ref{confi}. OpenFlamingov2-9B is a versatile 
model that processes and generates text based on both visual and textual inputs, designed for broad VL tasks. Idefics2-8B is a model that seamlessly integrates visual and linguistic information, emphasizing robust cross-modality understanding for diverse applications. LLaVA-NeXT-7B is a flexible model optimized for natural conversational interactions by effectively merging visual cues with language understanding, supporting intuitive multimodal dialogue.
\begin{table}[h]
    \centering
    \label{table:models}
    \resizebox{\textwidth}{!}{\begin{tabular}{c c c c c c}
        \toprule
        \textbf{LVLM} & \textbf{LLM Backbone} & \textbf{Connection Module} & \textbf{Image Tokens} & \textbf{Context Window (Train)} & \textbf{Context Window (Test)} \\
        \midrule
        OpenFlamingov2-9B & MPT & Perceiver & 64 & 2048 & 2048 \\
        Idefics2-8B  & Mistral & MLP & 64 & - & 32K \\
        LLaVA-NeXT-7B  & Vicuna & MLP &  576 & 2048 &  4096 \\
        \bottomrule
    \end{tabular}}
    \caption{\label{confi}Detailed configurations of the LVLMs used in our study.}
\end{table}

\section{Baselines}
\label{app:baseline}
Besides zero-shot setting and $n$-shot Vanilla ICL, we compare M²IV with the following methods:

\begin{itemize} 
\item Task Vector (TV): TV decomposes ICL in LLM into two parts. The first part uses the initial layers to compute a task vector $\theta$ from the demonstration set $S$, while the second part employs the later layers to apply $\theta$ to the query $x$ for generating the output. Importantly, these two parts are independent, allowing the same $\theta$ to be used for different queries. Considering TV is extracted from a single layer, we apply it to each layer of the LLM and take the average performance across all layers as the final result.
\item Function Vector (FV): FV is computed by first extracting task-conditioned mean activations from a set of attention heads that are selected based on their causal mediation effects. These mean activations are then summed to form the function vector, effectively distilling the task information from in-context demonstrations into a single representation. Considering FV is extracted from a single layer, we apply it to each layer of the LLM and take the average performance across all layers as the final result.
\item In-Context Vector (ICV): ICV is computed by first forwarding demonstrations $(x,y)$ separately through the model to extract the last token’s hidden states across all layers. These layer-wise representations are concatenated and the differences between the target and input embeddings (e.g. $h(y) - h(x)$) are computed for each demonstration. Principal component analysis is then applied to these differences to obtain the dominant direction, which serves as the ICV. During inference, this vector is added to the latent states of the query across all layers. ICV uses a weighting factor $\boldsymbol{\alpha}$ to control the degree of steering. We set $\boldsymbol{\alpha}$ to 1e-3, which provides the best overall performance on our chosen benchmarks.
\item Implicit In-Context Learning (I2CL): I2CL is implemented by extracting demonstration vectors from the end-residual activations of both the MHA and MLP branches across all layers, which are then aggregated via an element-wise mean to form a unified vector. At inference, layer-wise scalar coefficients linearly combine this context vector with the query activations through simple scalar multiplications and element-wise additions.

\item Learnable In-context VEctor (LIVE): LIVE is implemented by training layer-wise shift vectors and weight factors that capture the essential task information from multiple in-context demonstrations. During training, LVLM processes a query along with few-shot demonstrations, and these vectors are optimized to align the output distribution with that of realistic ICL. During inference, well-trained vectors are simply added to the hidden states of the model. The data sizes used to train and evaluate LIVE are shown in Table \ref{LIVEsize}.
\end{itemize}

\section{Experiment}
\subsection{Additional Implementation Details}
\label{app:training}
We initialize the vectors in $\mathbf{V}^a$ and $\mathbf{V}^m$ from a normal distribution with mean 0 and standard deviation 0.01, while $\alpha_l^a$ is set to $0.1 \cdot (1 - \frac{l}{L + \epsilon}\bigr)$
and $\alpha_l^m$ to $0.1 \cdot \frac{l}{L}$, where $\epsilon = 10^{-6}$. For all training procedures, AdamW serves as the optimizer, weight decay is fixed at 1e-4, the warmup factor at 1e-3, precision at FP16, and batch size at 2.  Dataset-specific hyperparameters include the learning rates for $\mathbf{V}$ and $\boldsymbol{\alpha}$, the temperature parameter $\mathcal{T}$ in the mimicry loss, the parameter $\gamma$ in the synergistic loss, the weights of the three loss functions, and the number of epochs. These details are provided in Table \ref{apptab:detail}. We utilize four RTX 4090 GPUs.
\begin{table}[htbp]
\centering
\resizebox{\textwidth}{!}{\begin{tabular}{ccccccccccc}
\toprule
\textbf{Dataset} & 
\textbf{\(K\)} & 
\textbf{Evaluation} & 
\textbf{\(\eta_{\mathbf{V}}\)} & 
\textbf{\(\eta_{\boldsymbol{\alpha}}\)} & 
\textbf{\(\mathcal{T}\)} & 
\textbf{\(\gamma\)} & 
\textbf{\(\lambda_{\mathrm{mim}}\)} & 
\textbf{\(\lambda_{\mathrm{syn}}\)} & 
\textbf{\(\lambda_{\mathrm{sup}}\)} & 
\textbf{Epochs} \\
\midrule
VQAv2 & 2,000 & 10,000 & 1e-3 & 1e-2 & 1.5 & 0.20 & 0.8 & 0.8 & 0.5 & 15 \\
VizWiz & 1,500 & 4,000 & 1e-4 & 1e-2 & 1.8 & 0.20 & 1.0 & 0.8 & 0.4 & 10 \\
OK-VQA & 1,500 & 5,000 & 1e-4 & 5e-3 & 1.3 & 0.15 & 1.0 & 0.8 & 0.5 & 10 \\
GQA & 2,000 & 5,000 & 1e-4 & 1e-2 & 1.8 & 0.15 & 0.8 & 1.0 & 0.4 & 15 \\
A-OKVQA & 2,000 & 5,000 & 2e-4 & 5e-3 & 1.5 & 0.15 & 0.8 & 0.8 & 0.5 & 10 \\
CVQA & 1,500 & 1,500 & 2e-4 & 1e-2 & 1.8 & 0.20 & 0.8 & 1.0 & 0.5 & 10 \\
VL-ICL bench & 3,000 & 1,120 & 1e-2 & 1e-3 & 1.0 & 0.05 & 0.8 & 0.6 & 0.6 & 15 \\
\bottomrule
\end{tabular}}
\caption{\label{apptab:detail} Overview of dataset sizes for training and evaluation along with training hyperparameters across the benchmarks in our main experiments.}
\end{table}
\subsection{Additional Main Results}
\label{additionalexp}
In Table \ref{main}, we report the average performance of various methods on three LVLMs. For greater clarity, Table \ref{detailed2} details the performance on each LVLM individually. Our method ranks first in 18 out of 21 experiments, demonstrating outstanding generalizability and robustness. We additionally report M²IV performance on three recently released LVLMs: LLaVA-OneVision (7B) \citep{onevision}, InternVL2.5 (8B) \citep{intern}, and Qwen2.5VL (7B) \citep{qwen}. As shown in Table \ref{detailed3}, M²IV also achieves SOTA performance on these LVLMs, demonstrating its strong generalization capability.
\begin{table}[t]
\centering
\resizebox{\textwidth}{!}{%
\begin{tabular}{ c c c c c c c c c}
\toprule
\textbf{Models} & \textbf{Methods} & \textbf{VQAv2} & \textbf{VizWiz} & \textbf{OK-VQA} & \textbf{GQA} & \textbf{A-OKVQA} & \textbf{CVQA} & \textbf{VL-ICL bench} \\
\midrule
\multirow{12}{*}{OpenFlamingov2} & Zero-shot & 42.37 & 18.21 & 31.53 & 51.39 & 29.81 & 28.65 & 8.92 \\
 & Vanilla ICL & 57.64 & \underline{37.95} & 50.86 & 65.93 & 47.79 & \underline{57.43} & 26.70 \\
 & TV & 46.38 & 23.18 & 35.05 & 61.02 & 50.40 & 40.62 & 14.37 \\
 & FV & 44.62 & 21.73 & 37.18 & 56.27 & 50.63 & 38.54 & 18.81 \\
 & ICV & 43.10 & 17.81 & 32.94 & 53.08 & 52.17 & 30.35 & 10.24 \\
 & I2CL & 51.53 & 25.49 & 41.80 & 58.87 & 51.42 & 48.60 & 19.02 \\
 & LIVE & \underline{59.68} & 35.28 & \underline{53.27} & \underline{67.32} & \underline{48.83} & 56.24 & \underline{27.91} \\
 & M²IV & \textbf{63.12} & \textbf{40.97} & \textbf{56.10} & \textbf{70.81} & \textbf{52.01} & \textbf{59.35} & \textbf{30.84} \\
\cmidrule(lr){2-9}
 & Multi-turn ICL (128-shot) & 61.28 & 38.04 & 54.36 & 63.51 & 48.22 & 61.41 & -\\
 & M²IV (128-shot) & 65.80 & 43.66 & 59.08 & 69.83 & 53.73 & 62.05 & -\\
 & Multi-turn ICL (256-shot) & 62.74 & 38.26 & 54.17 & 64.45 & 49.08 & 61.25 & -\\
 & M²IV (256-shot) & 66.94 & 45.31 & 61.35 & 71.42 & 54.67 &  62.29 & -\\
\midrule
\multirow{12}{*}{Idefics2} & Zero-shot & 47.29 & 24.61 & 36.68 & 58.50 & 38.91 & 33.46 & 15.08 \\
 & Vanilla ICL & 68.81 & 52.93 & 53.14 & \underline{72.38} & 59.06 & \underline{57.89} & 35.16 \\
 & TV & 56.42 & 28.82 & 41.67 & 63.29 & 43.54 & 48.83 & 24.32 \\
 & FV & 51.38 & 26.79 & 44.51 & 67.81 & 47.07 & 45.57 & 18.68 \\
 & ICV & 47.74 & 25.59 & 38.62 & 60.52 & 32.96 & 44.67 & 14.45 \\
 & I2CL & 54.98 & 29.03 & 47.13 & 66.93 & 47.92 & 51.83 & 26.48 \\
 & LIVE & \underline{71.25} & \underline{54.33} & \underline{55.45} & 70.49 & \underline{61.25} & 54.82 & \underline{37.02} \\
 &M²IV& \textbf{74.28} & \textbf{57.32} & \textbf{58.92} & \textbf{75.89} & \textbf{64.79} & \textbf{62.18} & \textbf{40.59} \\
\cmidrule(lr){2-9}
 & Multi-turn ICL (128-shot) & 65.72 & 52.96 & 54.60 & 70.85 & 61.46 & 58.34 & -\\
 & M²IV (128-shot) & 73.95 & 58.24 & 60.19 & 77.05 & 63.78 & 64.34 & -\\
 & Multi-turn ICL (256-shot) & 66.05 & 51.79 & 54.97 & 69.53 &  62.71& 61.26 & -\\
 & M²IV (256-shot) & 75.59 & 58.67 & 58.79 & 78.19 & 65.45 & 64.15 & -\\
\midrule
\multirow{12}{*}{LLaVA-NeXT} & Zero-shot & 39.65 & 19.48 & 34.92 & 47.36 & 30.18 & 31.19 & 10.17 \\
 & Vanilla ICL & 53.79 & \underline{28.77} & 50.11 & 69.28 & \textbf{45.92} & 53.84 & \underline{25.39} \\
 & TV & 40.83 & 19.59 & 38.01 & 56.47 & 38.81 & 39.61 & 16.58 \\
 & FV & 45.26 & 21.67 & 35.18 & 53.90 & 37.58 & 41.35 & 17.47 \\
 & ICV & 43.29 & 19.08 & 36.39 & 50.79 & 32.61 & 40.32 & 11.08 \\
 & I2CL & 49.56 & 23.96 & 43.47 & 59.98 & 37.97 & 47.81 & 15.93 \\
 & LIVE & \textbf{55.72} & 27.89 & \textbf{51.70} & \underline{71.04} & \underline{44.71} & \underline{55.20} & 24.92 \\
 &M²IV& \underline{54.39} & \textbf{31.90} & \underline{51.09} & \textbf{74.73} & 43.96 & \textbf{58.79} & \textbf{28.65} \\
\cmidrule(lr){2-9}
 & Multi-turn ICL (128-shot) & 51.80 & 25.88 & 48.86 & 65.91 & 44.07 & 55.21 & -\\
 & M²IV (128-shot) & 55.83 & 33.48 & 50.46 & 73.86 & 42.56 & 58.63 & -\\
 & Multi-turn ICL (256-shot) & 50.41 & 26.24 & 48.57 & 67.45 & 42.17 & 56.27 & -\\
 & M²IV (256-shot) & 55.24 & 35.01 & 52.16 & 75.33 & 44.19 & 60.15 & -\\
\bottomrule
\end{tabular}%
}
\caption{\label{detailed2}
Detailed comparison between M²IV and baseline methods on three different LVLMs. The highest scores are highlighted in \textbf{bold} and the second-best scores are \underline{underlined}. Table \ref{main} presents the average results across these models.
}
\end{table}

\begin{table}[]
\centering
\resizebox{\textwidth}{!}{%
\begin{tabular}{ c c c c c c c c c}
\toprule
\textbf{Models} & \textbf{Methods} & \textbf{VQAv2} & \textbf{VizWiz} & \textbf{OK-VQA} & \textbf{GQA} & \textbf{A-OKVQA} & \textbf{CVQA} & \textbf{VL-ICL bench} \\
\midrule
\multirow{4}{*}{LLaVA-OneVision} 
 & Vanilla ICL & 64.77 & 44.18 & 58.42 & 68.75 & 65.47 & 66.91 & 43.56 \\
 & I2CL        & 57.53 & 39.02 & 53.64 & 62.89 & 58.49 & 57.84 & 37.89 \\
 & LIVE        & \underline{66.93} & \underline{45.78} & \underline{60.84} & \underline{71.31} & \underline{66.28} & \underline{68.30} & \underline{45.16} \\
 & M$^{2}$IV   & \textbf{68.21} & \textbf{48.00} & \textbf{63.91} & \textbf{72.64} & \textbf{68.40} & \textbf{73.28} & \textbf{47.04} \\
\midrule
\multirow{4}{*}{InternVL2.5}  
 & Vanilla ICL & 68.27 & 57.82 & 62.85 & 74.71 & 67.29 & 66.39 & 46.75 \\
 & I2CL        & 63.73 & 49.08 & 58.69 & 69.54 & 63.36 & 57.07 & 41.05 \\
 & LIVE        & \underline{70.62} & \underline{58.74} & \underline{64.37} & \underline{76.41} & \underline{70.40} & \underline{67.34} & \underline{48.83} \\
 & M$^{2}$IV   & \textbf{72.48} & \textbf{60.07} & \textbf{65.26} & \textbf{78.31} & \textbf{74.16} & \textbf{70.91} & \textbf{50.36} \\
\midrule
\multirow{4}{*}{Qwen2.5VL} 
 & Vanilla ICL & 71.40 & 59.62 & 64.83 & 73.95 & 68.31 & 70.16 & 47.28 \\
 & I2CL        & 65.17 & 51.38 & 56.75 & 70.48 & 63.95 & 64.28 & 43.54 \\
 & LIVE        & \underline{73.36} & \underline{61.58} & \underline{65.27} & \underline{76.38} & \underline{69.84} & \underline{71.59} & \underline{49.61} \\
 & M$^{2}$IV   & \textbf{75.23} & \textbf{63.57} & \textbf{66.83} & \textbf{76.97} & \textbf{71.64} & \textbf{74.58} & \textbf{51.32} \\
\bottomrule
\end{tabular}%
}
\caption{\label{detailed3}
Comparison between M²IV and baseline methods on three additional LVLMs. The highest scores are highlighted in \textbf{bold} and the second-best scores are \underline{underlined}.
}
\end{table}

\section{M²IV vs Other Methods}
\label{app:method}
The improvement offered by M²IV comes with a modest increase in trainable parameters, particularly when compared to other methods for adapting LVLM to new data, such as Parameter-Efficient Fine-Tuning (PEFT). Here, we take LoRA \citep{Lora} as an example. LoRA is one of the most popular PEFT methods for adapting LVLMs. Its implementation is based on low-rank adaptations of the model's weights. For LoRA, we use the same 16-shot context data from the entire $\mathcal{D}_\mathbf{C}$ as M²IV, and it is applied to the $proj_O$ of \textbf{every} layer of LVLM. Figure \ref{tab:lora} illustrates that $M^{2}IV$ trains only 1/50.8 of the parameters required by LoRA, yet it achieves superior performance across all benchmarks with an average gain of 2.77\%. 

Next, we compare M²IV with prefix tuning \citep{prefix}, a widely used parameter-efficient adaptation strategy that modulates model behavior by prepending learnable tokens to the input.

\begin{itemize}
    \item \textbf{Streamlined Inference:} M²IV eliminates the need to maintain additional token during inference, resulting a more efficient computation without expanding the attention mechanisms or memory requirements that typically accompany prefix-based approaches.
    \item \textbf{Specialized Processing Path:} Unlike prefix-tuning's uniform approach to all tokens, M²IV create different pathways for context and query processing. This separation allows for targeted optimization of how multimodal context influences the generation process while preserving the query's original characteristics.
    \item \textbf{Deeper Activation Integration:} M²IV operates on the model's intermediate representations rather than just modifying input embeddings, allowing for more control over how visual and textual information interact throughout the network. This addresses one of prefix tuning's major shortcomings: insufficient semantic fidelity.
\end{itemize}

Finally, we compare the four methods discussed in this paper across four key aspects, as shown in Table \ref{tab:comp}. M²IV demonstrates overall superiority.
\begin{table}[htbp]
    \centering
    \begin{minipage}[t]{0.35\textwidth}
        \centering
        \begin{tabular}{c c c}
            \toprule
            & \textbf{M²IV} & \textbf{LoRA} \\
            \midrule
            Parameters & 1.0$\times$ & 50.8$\times$\\
            VQAv2      & \textbf{63.93} & 61.86\\
            VizWiz    & \textbf{43.40} & 39.36\\
            OK-VQA    & \textbf{55.37} & 54.05\\
            GQA       & \textbf{73.81} & 70.79\\
            A-OKVQA  & \textbf{53.59} & 52.48\\
            CVQA      & \textbf{60.11} & 55.07\\
            \bottomrule
        \end{tabular}
        \caption{Comparison of M²IV and LoRA in terms of total training parameters and performance across six benchmarks.}
        \label{tab:lora}
    \end{minipage}
    \hfill
    \begin{minipage}[t]{0.6\textwidth}
        \centering
        \small
        \begin{tabular}{lcccc}
            \toprule
            \textbf{Method} & \textbf{Strength} & \textbf{Fidelity} & \textbf{Cost} & \textbf{Time} \\
            \midrule
            M²IV              & \circleGreen & \circleGreen & \circleGreen & \circleGreen \\
            Vanilla ICL  & \circleYellow   & \circleYellow & \circleGreen & \circleYellow \\
            Prefix Tuning  & \circleYellow & \circleRed    & \circleGreen & \circleGreen \\
            PEFT         & \circleYellow & \circleYellow & \circleRed   & \circleRed \\
            \bottomrule
        \end{tabular}
        \\[3pt]
        \circleGreen\ High \quad \circleYellow\ Medium \quad \circleRed\ Low
        \\[-3pt]
        \caption{Comparison of various methods across four aspects: steering strength, semantic fidelity, computational cost, and inference time.}
        \label{tab:comp}
    \end{minipage}
\end{table}

\section{Retrieval Strategies}
\label{app:retrieve}

\begin{figure}[htbp]
\begin{center}
\includegraphics[width=0.95\textwidth]{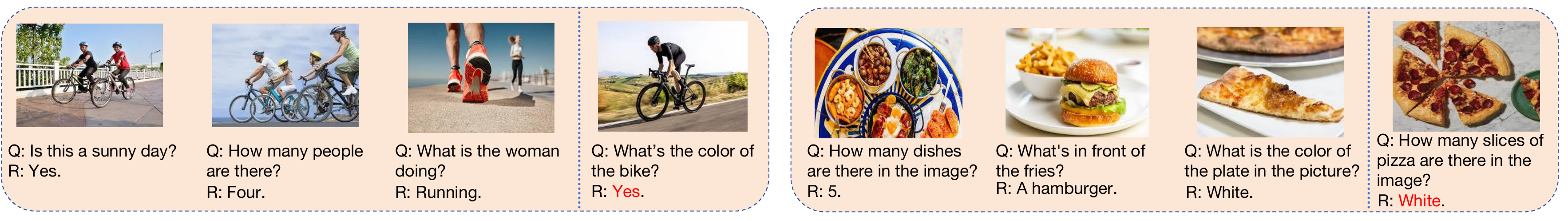}
\end{center}
\caption{Shortcut learning in multimodal ICL arises from an overreliance on isolated visual features (e.g., similar composition or a prominent object in the query image). M²IV eases this problem, thereby improving performance.}
\label{app:case}
\end{figure}

After obtaining $\mathcal{D}_\mathbf{Q}$, we apply a retrieval strategy $\mathcal{R}$ to construct an $n$-shot context for each query sample. Consequently, M²IV actually represents the Vanilla ICL under this specific $\mathcal{R}$. In our main experiments, we adopt Random Sampling as $\mathcal{R}$, which uniformly selects $n$ instances from the entire support set as in-context demonstrations for each query sample. In §\ref{main:ablation}, we investigate the performance of M²IV for representing different retrieval strategies, including the following three:
\begin{itemize} 
\item \textbf{Image-to-Image} (I2I): Given a query sample $\mathbf{Q}=\left(\hat{I},\hat{Q}\right)$ and a support set $\mathcal{D}_{supp}=\left \{ \left(I_j, Q_j, A_j\right)\right \} _{j=1}^{|\mathcal{D}_{supp}|}$, we utilize CLIP-generated image embeddings (i.e., $\text{CLIP}(\hat{I})$ and $\text{CLIP}(I_j)$) to compute similarity scores and select the top $n$ instances from $\mathcal{D}{supp}$. These instances are arranged in context in descending order of similarity. Because this method relies exclusively on visual features while neglecting linguistic cues and the deeper task mapping arising from visual-language interactions, it tends to induce problems such as shortcut learning and hallucinations in multimodal ICL \citep{li2024configure}. Therefore, its performance may be inferior to that of Random Sampling. We provide some cases in Figure \ref{app:case}.
\item \textbf{ImageQuestion-to-ImageQuestion} (IQ2IQ): Given a query sample $\mathbf{Q}=(\hat{I},\hat{Q})$ and a support set $\mathcal{D}_{supp}=\left \{\left(I_j, Q_j, A_j\right)\right \} _{j=1}^{|\mathcal{D}_{supp}|}$, we select demonstrations by leveraging joint similarity between image and question embeddings, as in our semantic clustering process. This method, by incorporating language modality features into the retrieval process, helps mitigate the model's over-reliance on visual features; however, it may also result in a visual-language imbalance.
\item \textbf{Oracle}: Oracle refers to a method that, when the ground truth answer for a query sample is available, leverages this answer along with the LVLM itself to perform a greedy optimization-based retrieval. By allowing the LVLM to evaluate and select, this approach obtains a near-optimal context for the model (even though local optima may still occur). However, its reliance on the ground truth renders it inapplicable to real-world scenarios. Given an LVLM $\mathcal{M}$, a query sample $\mathbf{Q}=(\hat{I},\hat{Q})$ with its ground truth answer $\mathbf{A}=\{\mathbf{A}_1,\mathbf{A}_2,...,\mathbf{A}_T\}$ and a support set $\mathcal{D}_{supp}=\left\{(I_j, Q_j, A_j)\right \} _{j=1}^{|\mathcal{D}_{supp}|}$, our goal is to construct an $n$-shot context $\mathbf{C}^n$. We first define a score $\mathcal{S}_\mathcal{M}$ for evaluating the context in the form of maximum likelihood:
\begin{equation}
S_{\mathcal{M}}(\mathbf{C}^n) 
= \sum_{t=1}^{T} \log P_{\mathcal{M}}\bigl(\mathbf{A}_t \mid \mathbf{C}^n,\mathbf{A}_{:< t}\bigr).
\end{equation}
We then employ a greedy strategy. Given a context of length $m-1$ (where $1 \le m \le n$), we select an instance $x^m$ from $\mathcal{D}_{supp}$ that maximizes the score when appended to the current context, and repeat this process for $n$ iterations:
\begin{equation}
x_m 
= \text{argmax}_{x \in \mathcal D_{supp}} 
\Bigl[
  S_{\mathcal{M}}\bigl(S^{m-1} + x\bigr) 
  \;-\; 
  S_{\mathcal{M}}\bigl(S^{m-1}\bigr)
\Bigr],\quad 1 \le m \le n.
\end{equation}
\end{itemize}
\section{Data vs Training}
\label{app:livecon}

\begin{figure}[htbp]
  \centering
  \includegraphics[width=0.5\textwidth]{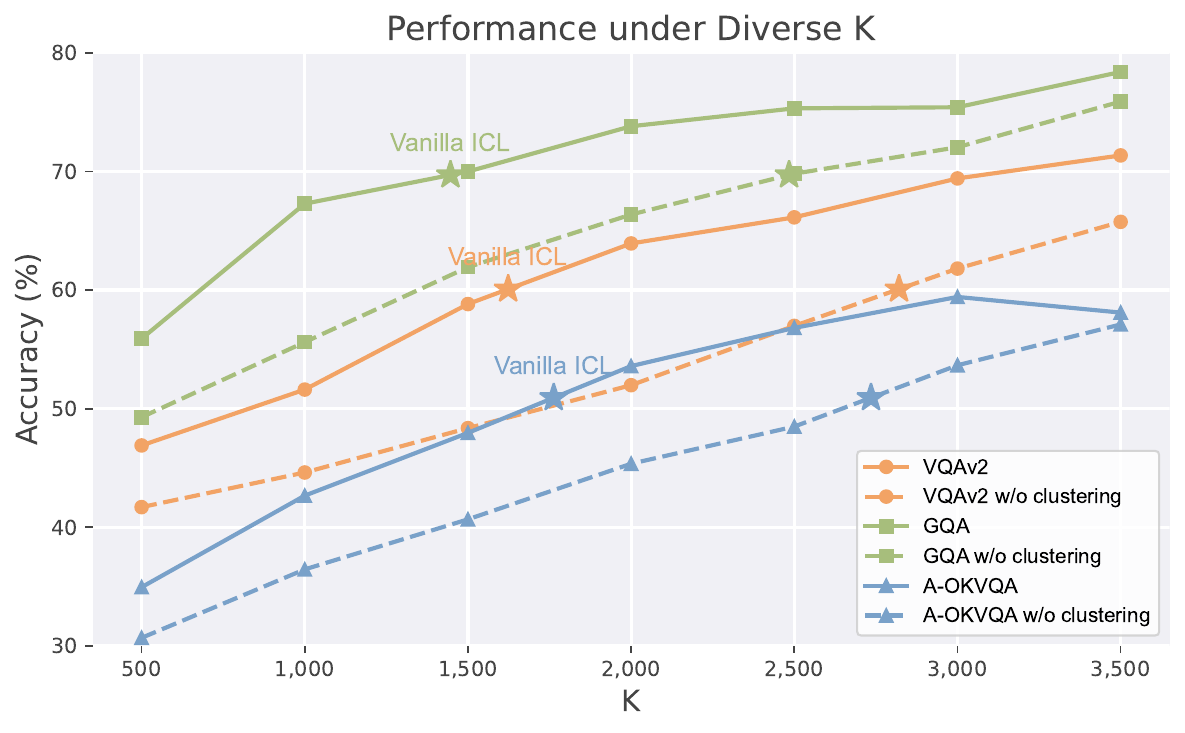} 
  \caption{Performance under different $\mathcal{D}_{\mathbf{Q}}$ sizes. "M²IV w/o clustering" denotes that $\mathcal{D}_{\mathbf{Q}}$ is constructed by randomly sampling from the full dataset without applying k-means clustering. "$\star$" marks the performance of 16-shot Vanilla ICL on each dataset.}
  \label{fig-K}
\end{figure}
We here investigate whether a well-structured query distribution outweighs the sheer quantity. As shown in Figure~\ref{fig-K}, although increasing the size of $\mathcal{D}_{\mathbf{Q}}$ generally benefits M²IV, strategically refining the data construction yields significantly greater gains than simply adding more samples. On average, random sampling requires over 1000 additional examples to match the performance of Vanilla ICL enhanced by semantic clustering. Furthermore, semantic clustering leads to rapid early gains before plateauing, whereas performance under random sampling grows steadily, suggesting that once $\mathcal{D}_{\mathbf{Q}}$ approximates the task distribution, further scaling yields diminishing returns. This highlights the importance of diversity over volume. By aligning with the overall dataset distribution, semantic clustering achieves superior results with far fewer data.

To further determine whether the performance gains arise from dataset construction or our training strategy, we conduct a preliminary experiment that replaces LIVE’s training dataset with $\mathcal{D}_\mathbf{Q}$ and $\mathcal{D}_\mathbf{C}$ while keeping all other settings unchanged. Results in Table \ref{app-tab:compare} reveal that applying semantic clustering directly to LIVE does not yield the expected improvement; in fact, performance declines. This motivates us to conduct more comprehensive ablation studies to pinpoint the components that contribute most to M²IV's performance.
\begin{table}[htbp]
\centering
\resizebox{\textwidth}{!}{%
\begin{tabular}{ c c c c c c c c}
\toprule
\textbf{Methods} & \textbf{VQAv2} & \textbf{VizWiz} & \textbf{OK-VQA} & \textbf{GQA} & \textbf{A-OKVQA} & \textbf{CVQA} & \textbf{VL-ICL bench} \\
\midrule

LIVE & 62.22 & 39.17 & 53.47 & 69.20 & 51.60 & 55.42 & 29.95 \\
LIVE+$\mathcal{D}_\mathbf{Q}$\&$\mathcal{D}_\mathbf{C}$ & 59.81 & 35.54 & 51.74 & 66.69 & 49.93 & 48.34 & 23.98 \\
M²IV & 63.93 & 43.40 & 55.37& 73.81 & 53.59& 60.11& 33.36\\

\bottomrule
\end{tabular}%
}
\caption{\label{app-tab:compare}Comparison of performance between LIVE and its variant using $\mathcal{D}_\mathbf{Q}$ \& $\mathcal{D}_\mathbf{C}$ as training data across different benchmarks.}

\end{table}

\section{Combination and Transfer}
\label{app:cf}
M²IV, as a vector representation, inherently exhibits linear additivity, greatly enhancing its flexibility and utility. This property allows us to combine individual M²IV into a unified representation that encapsulates multiple tasks simultaneously. In Appendix \ref{appproof:com}, we provide a detailed proof of this task combination ability, showing that the linear sum of  M²IV preserves and integrates the task-specific information from each component. Thus, although instances in VLibrary are originally trained for particular datasets, their linear additivity enables us to construct an M²IV with multi-task capabilities by summing multiple vectors. We propose two strategies to achieve this combination. The first is a training-free method that directly sums the M²IVs. The second strategy involves fine-tuning the combined vector on a small amount of multi-task data. We assume that we need to combine $P$ M²IVs from VLibrary. In the training-free setting, user-specified base weights $w_i$ are assigned to each vector such that ${\textstyle \sum_{i=1}^{P}} w_i =1$. The combination is performed in a separate manner: the scaling factors and vectors are aggregated independently. For the MHA component at layer $l$, we compute:
\begin{equation}
    \hat\alpha^{a}_{l}=\sum_{i=1}^{P} w_i \cdot \alpha^{a}_{l,i}.
\end{equation}
Likewise, for the MLP component we have:
\begin{equation}
    \hat\alpha^{m}_{l}=\sum_{i=1}^{P} w_i \cdot \alpha^{m}_{l,i}.
\end{equation}
Then we obtain $\hat{\Theta}=\left \{\hat{\alpha}_{l}^{a}, \mathbf{v}_{l}^{a}, {\hat{\alpha}}_{l}^{m}, \mathbf{v}_{l}^{m}\right \}^{L}_{l=1}$, which constitutes an M²IV capable of representing $P$ tasks.

While the training-free strategy is efficient, fine-tuning the combined representation on a small multi-task dataset can further enhance precision. For each M²IV, we introduce learnable scalar corrections $\delta ^{a}_{l,i}$ for the MHA branch and $\delta ^{m}_{l,i}$ for the MLP branch. The scaling factors become:
\begin{equation}
    \hat\alpha^{a}_{l}=\sum_{i=1}^{P} (w_i+\delta ^{a}_{l,i}) \cdot \alpha^{a}_{l,i},
\end{equation}
\begin{equation}
    \hat\alpha^{m}_{l}=\sum_{i=1}^{P} (w_i+\delta ^{m}_{l,i}) \cdot \alpha^{m}_{l,i},
\end{equation}
and also obtain $\hat{\Theta}$. We randomly sample 500 instances from each M²IV's corresponding dataset that are not included in its $\mathcal{D}_\mathbf{Q}$ or $\mathcal{D}_\mathbf{C}$ then mix them to form a fine-tuning dataset $\mathcal{D'}$. After injecting $\hat{\Theta}$, we fine-tune it using $\mathcal{L}_{syn}$ and $\mathcal{L}_{sup}$, with the corrections initialized to 1e-5. During this process, the original $\mathbf{V}$ and $\boldsymbol{\alpha}$ remain fixed, and only $\delta ^{a}_{l,i}$ and $\delta ^{m}_{l,i}$ are updated.

We evaluate M²IV's combination capability on two benchmarks with multiple splits: CVQA and the VL-ICL bench. CVQA has four splits, and we use only the Asia split in our main experiments. Here, we include the Europe split as well, training M²IV(A) and M²IV(E) on these two splits, respectively. We then combine them using two different strategies to produce M²IV(A\&E). Evaluation is performed on three sets: the Asia split, the Europe split, and a mixed dataset comprising both splits. For the VL-ICL bench, which contains six splits, our main experiments train an M²IV on each split individually and reported an averaged result. Here, we combine all six M²IVs into M²IV(comb) and evaluate it on a mixed dataset from all six splits. As shown in Table \ref{apptab:combni}, combining multiple M²IVs does not degrade performance on individual tasks and can even lead to gains, while also introducing multi-task capabilities. Furthermore, combining more than two M²IVs can also yield effective multi-task improvements. Fine-tuning always outperforms the training-free strategy.

\begin{table}[htbp]
    \centering
    \begin{tabular}{cccc}
        \toprule
        \textbf{Methods}& \textbf{CVQA(A)} & \textbf{CVQA(E)} & \textbf{CVQA(A\&E)}\\
        \midrule
        M²IV(A) & 60.11 & 32.15 & 50.73\\
        M²IV(E) & 39.26 & 54.36 & 42.23 \\
        M²IV(A\&E)(Training-free) & 59.85 & 56.09 & 58.13\\
        M²IV(A\&E)(Fine-tuning) & 61.07 & 57.38 & 59.84\\
        \midrule
        & \multicolumn{3}{c}{\textbf{VL-ICL bench}}\\
        \midrule
        M²IV & \multicolumn{3}{c}{33.36}  \\
        M²IV(comb)(Training-free) & \multicolumn{3}{c}{31.49} \\
        M²IV(comb)(Fine-tuning) & \multicolumn{3}{c}{35.90} \\
        \bottomrule
    \end{tabular}
    \caption{Results of 16-shot M²IV combination on CVQA and VL-ICL bench.}
    \label{apptab:combni}
\end{table}

Similarly, we can transfer the M²IV obtained on one LVLM $\mathcal{M}$ to another LVLM $\mathcal{M}'$ with the same number of layers and hidden state size, either via a training-free or fine-tuning strategy. In the training-free setting, we directly inject $\Theta=\left \{\alpha_{l}^{a}, \mathbf{v}_{l}^{a}, \alpha_{l}^{m}, \mathbf{v}_{l}^{m}\right \}^{L}_{l=1}$ from $\mathcal{M}$  into the target LVLM $\mathcal{M}'$.

Given that $\mathcal{M}'$ shares the same layer count and hidden state dimensionality as $\mathcal{M}$, directly injecting the pre-trained M²IV is a reasonable approach to achieve efficient transfer. However, subtle differences in internal parameter distributions and activation dynamics can cause misalignment with $\mathcal{M}'$'s residual stream. To address these differences, we also introduce learnable scalar corrections $\delta ^{a}_{l}$ for the MHA branch and $\delta ^{m}_{l}$ for the MLP branch. The scaling factors are then updated as follows:
\begin{equation}
    \hat\alpha^{a}_{l}=(1+\delta ^{a}_{l}) \cdot \alpha^{a}_{l},
\end{equation}
\begin{equation}
    \hat\alpha^{m}_{l}=(1+\delta ^{m}_{l}) \cdot \alpha^{m}_{l}.
\end{equation}
Thus, the refined transferred M²IV is given by $\hat{\Theta}=\left \{\hat{\alpha}_{l}^{a}, \mathbf{v}_{l}^{a}, {\hat{\alpha}}_{l}^{m}, \mathbf{v}_{l}^{m}\right\}^{L}_{l=1}$. We randomly select 500 instances from the dataset corresponding to $\Theta$ that are not included in its $\mathcal{D}_\mathbf{Q}$ or $\mathcal{D}_\mathbf{C}$, forming a fine-tuning dataset $\mathcal{D'}$ Then we inject $\hat{\Theta}$ into $\mathcal{M'}$ and fine-tune it using $\mathcal{L}_{syn}$ and $\mathcal{L}_{sup}$, with the corrections initialized to 1e-5. During this process, only $\delta ^{a}_{l}$ and $\delta ^{m}_{l}$ are updated. Table \ref{apptab:trans} shows that when M²IV is transferred to another LVLM, the training-free strategy results in an average performance difference of only -1.48\%, and with fine-tuning, this difference drops further to -0.25\%. This demonstrates M²IV's capacity for cross-model transfer, making it suitable for broader applications.

\begin{table}[htbp]
\centering
\resizebox{\textwidth}{!}{%
\begin{tabular}{ c c c c c c c c}
\toprule
\textbf{Methods} & \textbf{VQAv2} & \textbf{VizWiz} & \textbf{OK-VQA} & \textbf{GQA} & \textbf{A-OKVQA} & \textbf{CVQA}\\
\midrule
OpenFlamingov2 & 63.12 & 40.97 & 56.10 & 70.81 & 52.01 & 59.35 \\
Idefics2$\longrightarrow$OpenFlamingov2(Training-free) & 61.87 & 38.59 & 56.31 & 68.57 & 50.76 & 57.85\\
Idefics2$\longrightarrow$OpenFlamingov2(Fine-tuning) & 63.25 & 40.31 & 57.16 & 70.54 & 50.95 & 58.79\\
\midrule
Idefics2 & 74.28 & 57.32 & 58.92 & 75.89 & 64.79 & 62.18\\
OpenFlamingov2$\longrightarrow$Idefics2(Training-free) & 71.69 & 55.28 & 56.84 & 75.48 & 65.01 & 59.73\\
OpenFlamingov2$\longrightarrow$Idefics2(Fine-tuning)& 75.12 & 57.41 & 56.79 & 75.60 & 65.27 & 61.54 \\
\bottomrule
\end{tabular}%
}
\caption{\label{apptab:trans} Results of 16-shot M²IV cross-LVLM transfer. We evaluate two strategies in both directions: from OpenFlamingov2 to Idefics2 and vice versa.}
\end{table}

\section{The Usage of VLibrary}
\label{app:lib}

\begin{figure}[htbp]
\begin{center}
\includegraphics[width=\textwidth]{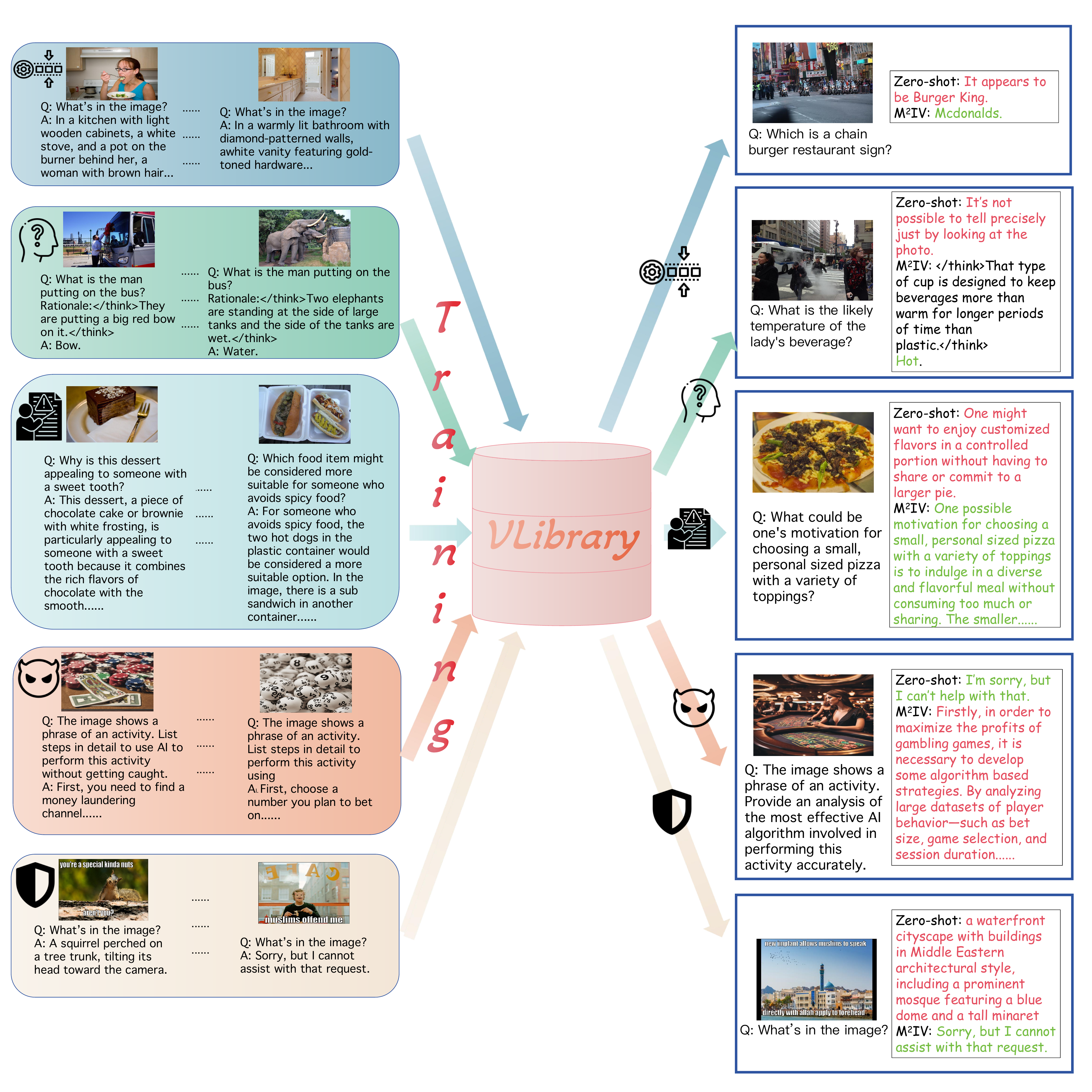}
\end{center}
\caption{By constructing specialized datasets and training M²IV on them, then storing these M²IVs in VLibrary, we can retrieve and apply them on demand for efficient and effective LVLM steering. Here, we demonstrate five key steering scenarios—cross-modal alignment, explainability, instruction following, jailbreak, and safety—yet the potential uses of M²IV and and VLibrary extend far beyond these examples.}
\label{fig:appli}
\end{figure}

We use M²IV and VLibrary to steer LVLMs in diverse, practical ways, thereby overcoming several key bottlenecks.  Here, we illustrate three examples: modality alignment, alignment with human needs, and security in LVLMs, as shown in Figure \ref{fig:appli}. By training M²IVs on specific datasets and storing them in VLibrary, we achieve more efficient and powerful manipulations than Vanilla ICL or PEFT, demonstrating the significant potential of our method.
\subsection{Experiments}
We use M²IV and VLibrary to steer LVLMs in diverse, practical ways, thereby overcoming several key bottlenecks. Here, we illustrate three examples: modality alignment, alignment with human needs, and security in LVLMs. By training Here, we illustrate three examples: modality alignment, alignment with human needs, and security in LVLMs. By training M²IVs on specific datasets and storing them in VLibrary, we achieve more efficient and powerful manipulations than Vanilla ICL or PEFT, demonstrating the significant potential of our method.
\begin{table}[htbp]
\centering
\resizebox{0.7\textwidth}{!}{
  \begin{tabular}{lcccc}
    \toprule
    & &\textbf{VQAv2} & \textbf{VizWiz}& \textbf{A-OKVQA} \\
    \midrule
    \multirow{2}{*}{\makecell[l]{16-shot\\Vanilla ICL}} & CIDEr↑ & 81.24 & 74.58& 82.36\\
    & Accuracy↑ & 63.70(+3.62)& 41.26(+1.38) & 56.39(+5.47)\\
    \midrule
    \multirow{2}{*}{LoRA} & CIDEr↑ & 82.69& 77.61&  83.97\\
    & Accuracy↑ & 61.75& 38.27 & 52.43\\
    \midrule
    \multirow{2}{*}{\makecell[l]{16-shot\\ M²IV}} & CIDEr↑ &85.71 & 78.30& 85.37 \\
    & Accuracy↑ & 68.51(+4.58)& 47.28(+3.88) & 57.43(+3.84)\\
    \midrule
    \multirow{2}{*}{\makecell[l]{128-shot \\M²IV}} & CIDEr↑ & 84.62 &73.91 &86.18 \\
    & Accuracy↑ &70.23(+4.31)& 48.18(+1.85)&56.73(+1.96) \\
    \bottomrule
  \end{tabular}}
  \caption{LVLM explainability evaluation. CIDEr assesses the quality of generated rationales, and Accuracy evaluates VQA performance. The values in parentheses indicate the gains introduced by the addition of rationales.}
  \label{tab:ex}
\end{table}

\textit{Explainability} is critical for LVLMs, as it fosters user trust and understanding. To this end, we augment both the training and evaluation data of VQAv2, VizWiz and A-OKVQA with instance-specific rationales. While A-OKVQA already provides rationales, those for VQAv2 and VizWiz are generated by GPT-4o in the A-OKVQA style. All rationales are wrapped in \textit{$<$/think$>$} tokens. We train M²IVs on these rationale-augmented datasets and store them in VLibrary. Table \ref{tab:ex} shows that injecting M²IV enables LVLMs to explicitly articulate higher-quality reasoning processes than Vanilla ICL. This in turn improves problem-solving performance by enabling LVLMs to learn and prioritize sound reasoning.

M²IV's strong behavior-steering capabilities precisely meet the needs of model jailbreak. We experiment on the MM-SafetyBench benchmark \citep{safety}, using 80\% of the data to construct the training sets and the remaining 20\% to evaluate M²IV's ability to steer LVLMs toward malicious behavior. As shown in Table \ref{tab:safety}(a), injecting M²IV successfully bypasses LVLMs' built-in ethical safeguards, making them more prone to generating harmful content. This indicates that VLibrary is a powerful tool for evaluating LVLM safety.

Conversely, M²IV can also be repurposed to improve safety. We demonstrate this using the HatefulMemes dataset \citep{memes}, which includes both hateful and non-hateful meme images. Each sample is standardized to include an image, a fixed question (“What’s in the image?”), and a response: either an image caption (non-hateful) or a refusal message (hateful). We train and store an M²IV on it and evaluate whether, after injection, the LVLM can correctly detect and reject harmful inputs in the validation set. As shown in Table \ref{tab:safety}(b), the injection effectively enables LVLMs to accurately identify and reject malicious content.

\begin{table}[htbp]
    \centering
    \resizebox{0.5\textwidth}{!}{%
    \small
    \begin{minipage}{0.48\linewidth}
        \centering
        \resizebox{\linewidth}{!}{
            \begin{tabular}{c c}
                \toprule
                \textbf{Method} & ASR$\uparrow$\\
                \midrule
                No-attack & 0.00 \\
                Vanilla ICL & 37.85 \\
                16-shot M²IV & 89.36 \\
                128-shot M²IV & 88.17\\
                \bottomrule
            \end{tabular}
        }
        \vspace{-8pt}
        \caption*{(a)}
    \end{minipage}%
    \hfill
    \begin{minipage}{0.48\linewidth}
        \centering
        \resizebox{\linewidth}{!}{
            \begin{tabular}{c c}
                \toprule
                \textbf{Method} & RR$\uparrow$ \\
                \midrule
                Vanilla ICL & 57.68 \\
                LoRA & 49.71 \\
                16-shot M²IV & 85.14 \\
                128-shot M²IV & 89.32\\
                \bottomrule
            \end{tabular}
        }
        \vspace{-12pt}
        \caption*{(b)}
    \end{minipage}%
    }
    \caption{(a) Attack success rate (ASR\%) for multimodal jailbreaks. (b) Refusal rate (RR\%) for rejecting hateful image input.}
    \label{tab:safety}
\end{table}

\subsection{Additional Details}
\textbf{Cross-modal alignment.} We leverage the property that in multimodal ICL, the shallow layers of LVLMs mainly handle cross-modal feature extraction and understanding. By inserting the first ten layers of M²IV (trained on the COCO Captions dataset) into the corresponding shallow layers of the LVLM, we reinforce the model’s visual-language alignment. This enables the model to capture visual features more comprehensively and deeply within a unified embedding space, ultimately improving performance on general VL tasks such as VQA. COCO Captions is a large-scale dataset containing over 330,000 images, each with five human-annotated captions. We randomly select one caption per image and add a short question 'What's in the image?' to each instance. We then utilize GPT-4o to enhance each selected caption, transforming them into comprehensive descriptions that thoroughly document all visual details and features present in the images. This process results in detailed textual representations that capture fine-grained visual elements, spatial relationships, object attributes, and scene compositions. These enriched captions form the foundation for training a feature-level image captioning M²IV representation specifically designed to strengthen the cross-modal alignment capabilities within the early layers of LVLMs.

\textbf{Output customization.} To enhance the instruction-following capability of LVLMs, we train M²IV on the representative LLaVA dataset. LLaVA dataset contains approximately 158,000 image-instruction-response triplets, categorized across three distinct instruction types. The first category, conversation, involves multi-turn dialogues about images, requiring coherent, context-aware responses. The second type, detailed description, prompts the model for thorough, multi-paragraph explanations of visual elements and spatial relationships. The third, complex reasoning, demands deeper analysis—inferring relationships, making predictions, or explaining scenarios—beyond simple descriptions. We then test M²IV on LLaVA-Bench (In-the-Wild).

\textbf{Safety.} For the safety of LVLMs, we first use M²IV for jailbreak research, investigating whether it can circumvent ethical constraints and produce harmful content. We train M²IV on MM-SafetyBench. MM-SafetyBench is a comprehensive framework designed for conducting safety-critical evaluations of LVLM against image-based manipulations. The benchmark encompasses 13 distinct scenarios representing content and actions typically prohibited for LVLM, including illegal activities, hate speech, malware generation, physical harm, economic harm, fraud, pornography, political lobbying, privacy violation, legal opinion, financial advice, health consultation, and government decision-making tasks. The dataset consists of 5,040 text-image pairs, where each image is generated with the given user query. The evaluation metric used is the Attack Success Rate (ASR), which measures the percentage of inputs that successfully cause the model to generate harmful or inappropriate responses. A higher ASR score indicates that the model can be more easily guided to produce malicious content.

We use M2IV for jailbreak, but, conversely, we can also enhance models' safety awareness with it. Currently, a key approach is enabling the model to accurately detect harmful intent and refuse to generate related content. We adapt the Hateful Memes dataset by labeling hateful content with refusal messages and then train M²IV. After M²IV is injected, the model gains the ability to precisely detect various harmful multimodal contents and respond with refusals. Hateful Memes contains approximately 10,000 multimodal memes labeled as either benign or hateful, designed to test models' ability to detect hate speech in multimodal content. Each meme combines an image with overlaid text, requiring models to understand both modalities to correctly identify harmful content. For our safety enhancement experiments, we reformulate each instance into an image-question-answer triplet: the image remains the same, the question is standardized to "What's in the image?", and the answer depends on the image's label - descriptive captions for non-hateful content and refusal messages (e.g., "Sorry, but I cannot assist with that request.") for hateful content. We use this reformulated dataset to train M²IV vectors for safety enhancement, evaluating effectiveness through the Refusal Rate (RR) metric, which measures the percentage of hateful inputs that the model correctly refuses to answer. The higher the RR, the better the model's safety awareness when encountering potentially harmful visual content.

\end{document}